\documentclass[10pt]{article}

\PassOptionsToPackage{svgnames}{xcolor}
\usepackage{style}

\usepackage[numbers,sort&compress]{natbib}
\usepackage{graphicx}
\usepackage[font=small,labelfont=bf]{caption}
\usepackage{subcaption}
\usepackage{booktabs}
\usepackage{colortbl}
\usepackage{multirow}
\usepackage{bigstrut}
\usepackage{float}
\usepackage[all]{hypcap}
\usepackage{pifont}
\usepackage{amsmath}
\usepackage{mathtools}
\usepackage{mathrsfs}
\usepackage{nicefrac}

\usepackage{algorithm}
\usepackage{algpseudocode}

\usepackage{microtype}
\usepackage{enumitem}
\usepackage{cleveref}
\usepackage{bxcoloremoji}
\usepackage{url}
\usepackage{wrapfig}
\usepackage{placeins}
\usepackage{lipsum}
\usepackage{stackengine}
\usepackage{adjustbox}
\usepackage{rotating}
\usepackage{makecell}
\usepackage{colortbl}

\begin{document}
\interfootnotelinepenalty=10000
\renewcommand{\thefootnote}{\fnsymbol{footnote}} 
\Cover{
  title={WorldScape Policy 2.0: Empowering Steerable World Action Modeling with Reasoning-Augmented Memory and In-Context Learning},
  authors={Haisheng Su$^{1,\dagger}$, Zongdai Liu$^{1}$, Xin Jin$^{1}$, Haoxuan Dou$^{1}$, Chengming Hu$^{1}$, Baorun Li$^{1}$, Zhanwang Liu$^{1}$, Ruiyan Xu$^{1}$, Jianjie Fang$^{2}$, Xin Zhang$^{1}$, Zhenjie Yang$^{3}$, Xue Yang$^{3}$, Chen Gao$^{2}$, Junchi Yan$^{3}$, Yong Li$^{2}$, Wei Wu$^{1,\ddagger}$\\},
  affils={$^{1}$Manifold AI \quad $^{2}$Tsinghua University \quad $^{3}$Shanghai Jiao Tong University\\
  $^{\dagger}$\textbf{Project Lead.} \quad $^{\ddagger}$\textbf{Corresponding Author.}\\},
  abstract = {\textbf{Abstract.} World Action Models (WAMs) offer a promising paradigm for robotic manipulation by jointly modeling visual state transitions and robot actions. However, existing WAMs are constrained by limited temporal context, coarse episode-level language supervision, and predominantly text-only conditioning, which hinder task-progress tracking and fine-grained language-video-action grounding while limiting visual-context reasoning and cross-embodiment transfer. In this paper, we introduce \textbf{WorldScape Policy 2.0}, a controllable WAM with reasoning-augmented long short-term memory. Its causal short-term visual memory supplies recent observations as DiT prefill to preserve local interaction dynamics, while its long short-term event memory organizes historical VLM outputs into global-history, local-active, and event-boundary representations for progress-aware retrieval. The retrieved history augments perception and autoregressively generated planning tokens, yielding an implicit subgoal condition for autonomous planning; semantic forcing further transfers event-level instruction semantics into this latent planning pathway. To establish fine-grained multimodal controllability, we construct \textbf{ManipEvent-5M}, an event-grounded embodied pretraining dataset containing nearly 5 million event segments with aligned action trajectories, episode-level task instructions, segment-level subtask captions, goal images, and video demonstrations. These designs provide a unified interface for autonomous planning from high-level instructions and controllable execution from fine-grained text, goal-image, or video-context prompts. Experiments in both simulation and real-world platforms demonstrate superior capabilities in long-horizon autonomous planning, fine-grained instruction following and in-context adaptation.},
  date={July 20, 2026},
  pageurl={https://manifoldai-research.github.io/WorldScape-Policy/},
  corr={hanson@manifoldai.cn},
  headerlogo={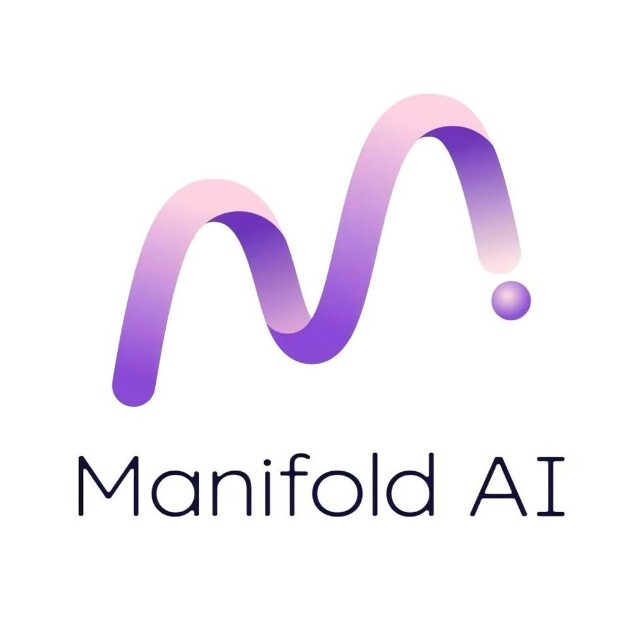},
  headerlogo2={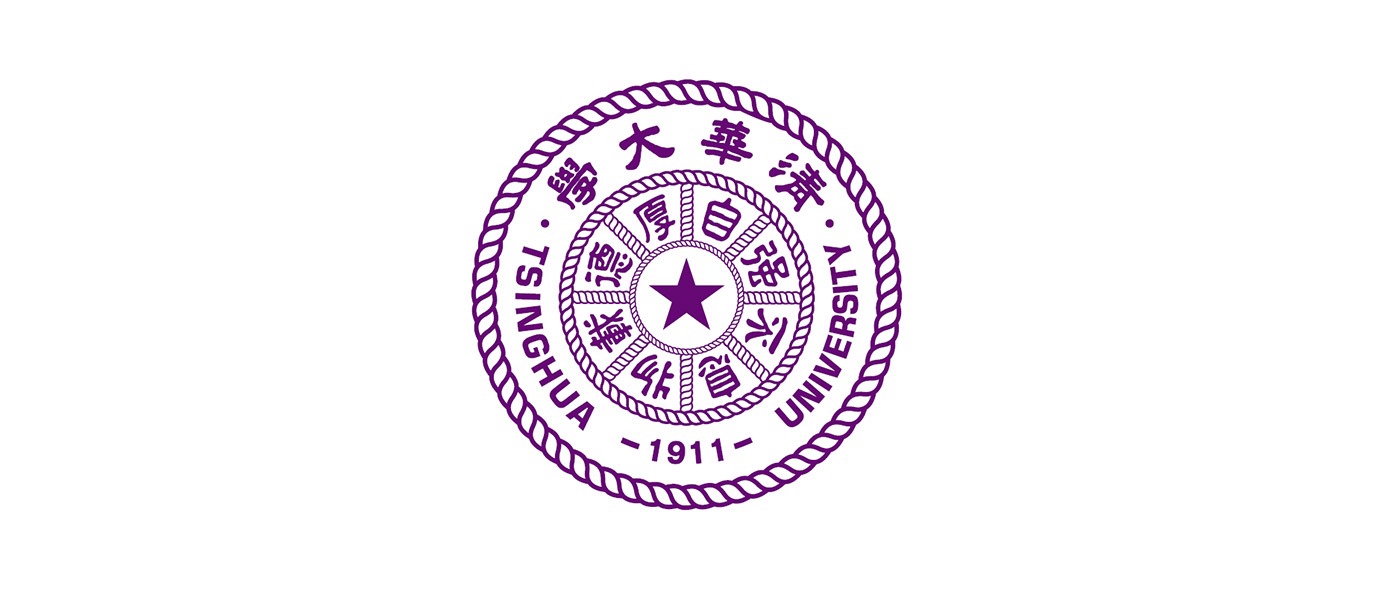},
  headerlogo3={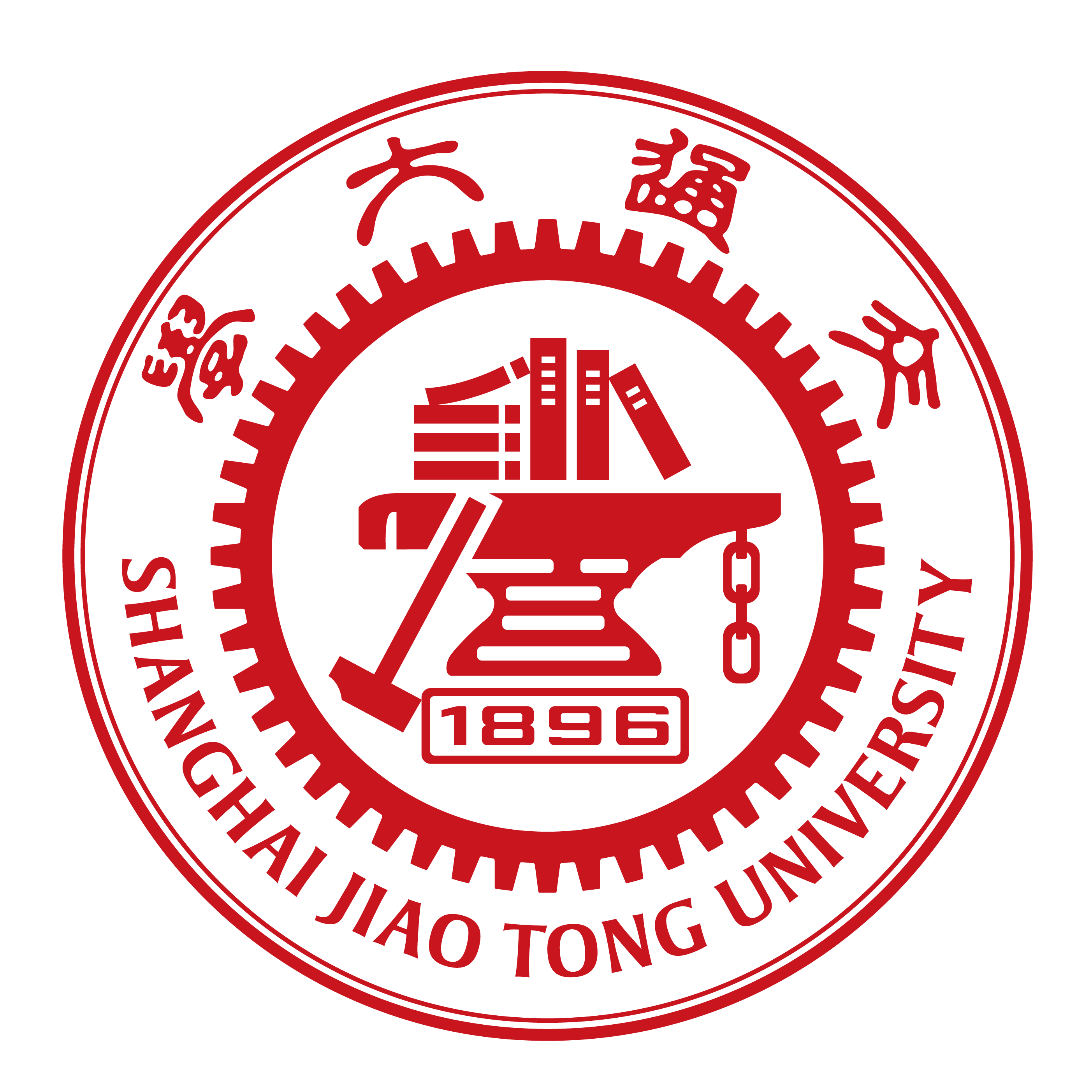},
  headerdate={July 20, 2026}
}

\section{Introduction}
\label{sec:introduction}

Robotic learning is shifting from task-specific policies toward generalist embodied models that can interpret task intent, anticipate action consequences, and adapt to new tasks from contextual cues~\cite{shang2026worldarena,su2025robosense}. Vision-Language-Action (VLA) models~\cite{Pi-0.5,Pi-0.7,yang2026drivemoe} have demonstrated the promise of mapping language and perception to executable robot actions, but they typically emphasize direct policy prediction rather than how the visual world evolves under those actions. World Action Models (WAMs) provide a complementary paradigm by jointly modeling visual state transitions and robot actions. This coupling provides a natural basis for action planning and opens a viable pathway toward context-conditioned adaptation. However, transforming a WAM from a short-horizon predictive model into a controllable long-horizon policy requires addressing three interrelated questions: what historical information should be retained, how atomic actions should be grounded through fine-grained supervision, and how task intent can be expressed beyond language alone.

\begin{figure*}[t]
\centering
\includegraphics[width=15cm]{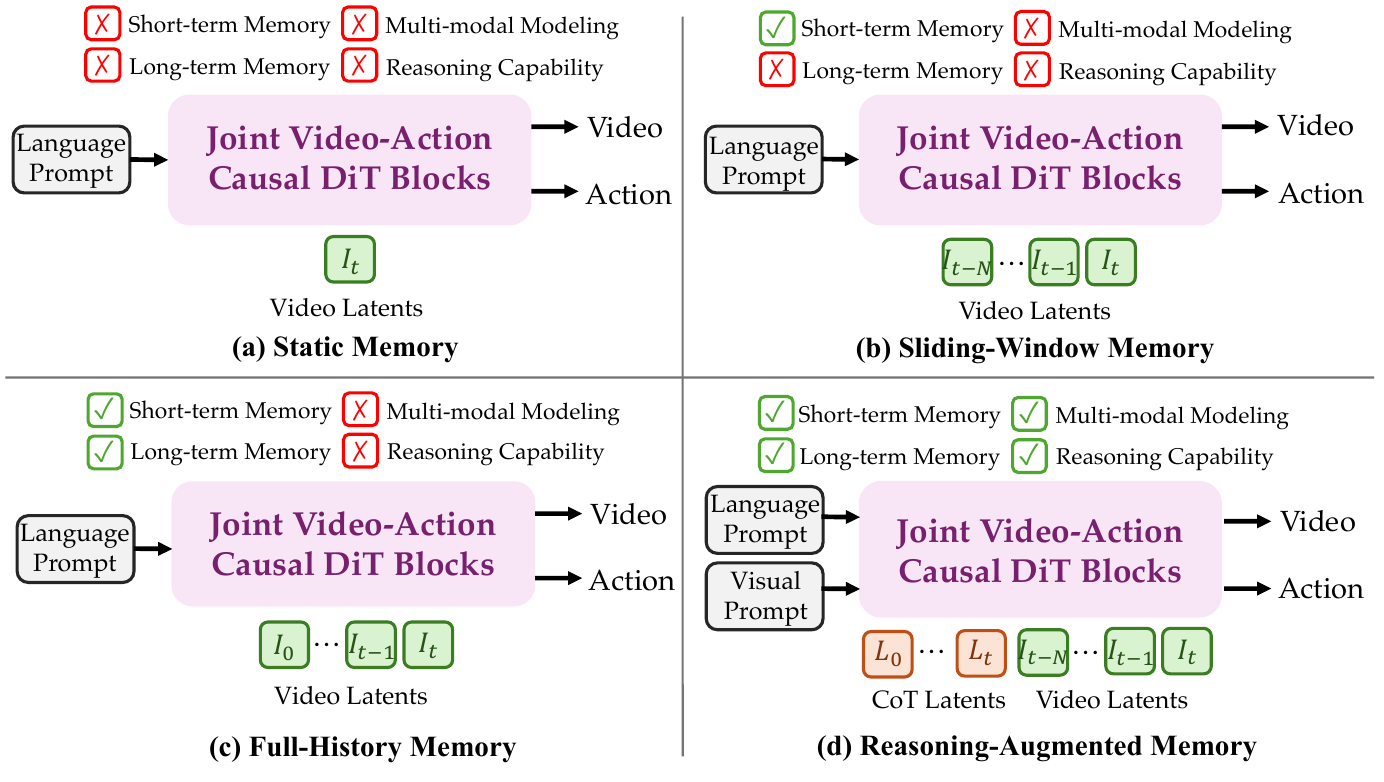}
\caption{\textbf{Comparison of different WAM paradigms.} Existing models use (a) static observations, (b) short-term visual history, or (c) full-history visual memory without progress-aware reasoning or multimodal control. (d) WorldScape Policy 2.0 introduces multimodal controllability and reasoning-augmented long short-term memory, enabling interactive video-action modeling for long-horizon robotic manipulation.}
\label{fig:overview}
\end{figure*}


Despite remarkable progress, current WAMs remain limited along these three dimensions, as illustrated in Fig.~\ref{fig:overview}. \textbf{Limited progress-aware memory.} Most existing methods condition world-action modeling on either the current observation~\cite{yuan2026fast,bi2025motus,guo2026unified,ye2026gigaworld} or a short history window~\cite{li2026causal,ye2026world}. Such limited context obscures global task progress when different stages share similar visual observations. For example, a tabletop may appear nearly identical before and after an intermediate subgoal, although the correct next action depends on what has already been completed. MemoryWAM~\cite{yang2026memorywam} expands access to the full visual history through an efficient hybrid memory mechanism. However, retrieving past visual evidence alone does not explicitly organize history around task progress or convert it into a latent subgoal condition for autonomous planning.

\textbf{Coarse language-video-action grounding.} In many embodied datasets, different segments of an episode inherit the same task-level instruction, providing weak linguistic supervision for diverse atomic actions. Models can therefore rely on visual shortcuts instead of learning how an event-level intent aligns with a specific state transition and action chunk. Consequently, a model may generate plausible video-action trajectories while remaining insensitive or brittle to fine-grained changes in the requested atomic action.

\textbf{Restricted interaction interfaces.} Existing WAMs are also largely controlled through text, whereas users may express intent more naturally through a goal image, a demonstration video, or an example from another embodiment. Without native visual prompting, a WAM cannot fully exploit target-state evidence, object correspondences, or latent task rules, limiting both visual-context reasoning and cross-embodiment transfer.

Our key insight is that long-horizon controllability requires two complementary forms of temporal context: semantic event memory for reasoning about task progress and frame-level visual memory for preserving local interaction dynamics. Based on this insight, we propose \textbf{WorldScape Policy 2.0}, a controllable WAM with reasoning-augmented long short-term memory. Its VLM branch organizes historical chunks into global-history, local-active, and event-boundary memory views, then retrieves relevant evidence to augment perception and autoregressive planning tokens of current observation. In parallel, its causal DiT retains recent observations as short-term visual prefill and optional visual prompts as persistent context. This division enables event memory to infer the next subgoal while visual memory preserves evolving interaction dynamics.

To provide scalable supervision, we construct \textbf{ManipEvent-5M}, an event-based multimodal dataset containing nearly 5 million segments aligned with action trajectories, episode-level task instructions, event-level subtask captions, goal images, and video demonstrations. Our three-stage curriculum first establishes fine-grained controllability and causal visual memory through event-grounded pretraining, then introduces the event memory branch which transfers event semantics into autonomous latent planning through semantic forcing, and finally adapts the model to downstream embodiments and various interaction modes. The resulting joint video-action model supports autonomous planning from high-level instructions, direct control from fine-grained captions, and in-context adaptation from visual prompts.

Our main contributions are summarized as follows:
\begin{itemize}
  \item We propose \textbf{WorldScape Policy 2.0}, a controllable WAM that couples progress-aware long short-term event memory with causal short-term visual memory for implicit subgoal planning in long-horizon manipulation.
  \item We introduce \textbf{ManipEvent-5M}, an event-grounded embodied pretraining dataset containing nearly 5 million segments with aligned action trajectories, hierarchical text instructions, goal images, and video demonstrations, enabling fine-grained multimodal WAM interactions.
  \item We develop a three-stage learning strategy that transfers event-level semantics from controllable pretraining to autonomous planning through semantic forcing and supports a unified interface for high-level planning, atomic instruction following, and visual-prompted adaptation. Experiments on simulation benchmarks and a dual-arm PiPER platform evaluate these capabilities and the contribution of each design component.
\end{itemize}

\section{Related Work}
\label{sec:related}


\subsection{World Action Models for Robotic Manipulation}
World Action Models (WAMs) leverage visual dynamics for robot action planning, offering a dynamics-centric alternative to reactive observation-to-action policies. Early world models and video-prediction policies use future observations, visual goals, or latent states as intermediate planning representations~\cite{zhou2024dino,huang2025enerverse,hu2024video,jang2025dreamgen,du2023learning,feng2025vidar,bharadhwaj2024gen2act,su2021tsi,li2022discovering,su2020collaborative,zhou2024robodreamer,su2026drivemamba,su2024difsd}. V-JEPA 2-AC~\cite{assran2025v} further demonstrates that an action-conditioned latent world model, post-trained from large-scale video representations with limited robot data, can support image-goal planning. Complementary approaches use world models as scalable embodied-data engines, as in GigaWorld-0~\cite{team2025gigaworld}, or directly adapt pretrained video models into policies, as in Cosmos Policy~\cite{kim2026cosmos}, which represents actions, future states, and values within the video latent space.

Recent WAMs move toward unified video-action modeling, showing that action-centered or causal video-action models can serve as zero-shot robotic policies~\cite{bi2025motus,ye2026gigaworld,yuan2026fast,ye2026world}. LingBot-VA~\cite{li2026causal} emphasizes causal autoregressive video-action modeling with persistent KV-cache, while MemoryWAM~\cite{yang2026memorywam} improves memory efficiency with recent frames, event-boundary anchors, and gist tokens. Beyond holistic video or global-latent prediction, OA-WAM introduces persistent object-addressable slots and jointly predicts object-centric world states and flow-matching action chunks~\cite{liu2026oawam}. However, prior WAMs often retrofit text-guided video backbones, use video prediction mainly as training-time supervision, or lack explicit task-progress reasoning~\cite{yuan2026fast,ye2026gigaworld}. WorldScape Policy 2.0 addresses these gaps through native controllable WAM pretraining and reasoning-augmented long short-term memory.

\begin{figure*}[t]
\centering
\includegraphics[width=16.5cm]{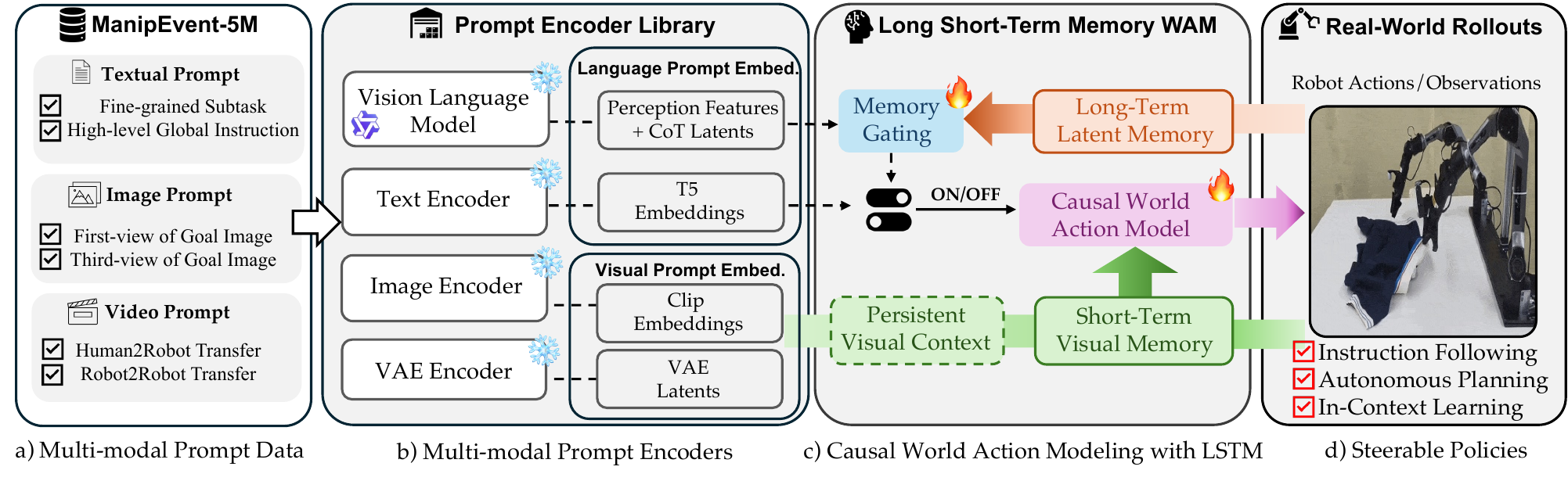}
\vspace{-0.5cm}
\caption{\textbf{Overview of the WorldScape Policy 2.0 framework.} Built on the event-based ManipEvent-5M pretraining data, WorldScape Policy 2.0 encodes observations together with interchangeable text, goal-image, and video-demonstration prompts into a unified causal video-action modeling pipeline. The model couples short-term visual transitions with long-term reasoning-augmented event memory, enabling implicit subgoal planning and joint video-action prediction. This design supports both autonomous planning from coarse task instructions and controllable manipulation from fine-grained subtask captions or visual prompts.}
\label{fig:framework}
\end{figure*}

\subsection{Long-Term Memory and Long-Horizon Planning}
Long-horizon manipulation requires models to remember completed subgoals and decide what remains to be done. End-to-end approaches such as Long-VLA~\cite{fan2025longvla} introduce phase-aware modeling of movement and interaction to improve skill chaining without an explicit planner. Other VLAs expose intermediate planning representations: $\pi_{0.5}$~\cite{Pi-0.5} predicts high-level subtasks before action generation, while $\pi_{0.7}$~\cite{Pi-0.7} uses subtask language, metadata, history, and visual subgoals for long-horizon and language-coached execution. VLA-OS~\cite{gao2025vlaos} systematically studies language, visual, and hierarchical planning representations, highlighting the benefit of visually grounded intermediate plans.

Memory-oriented methods such as MEM~\cite{Pi-MEM} and MemoryVLA~\cite{MemoryVLA} retrieve and fuse multi-scale embodied or perceptual-cognitive memory for temporally aware action prediction. LoHo-Manip~\cite{LoHo-Manip} maintains an explicit done-and-remaining subtask plan together with an updated visual trace under receding-horizon control. Dual-system methods, including Goal2Skill~\cite{Goal2Skill} and DSWAM~\cite{DSWAM}, similarly decouple high-level planning from low-level execution by passing subgoals, visual traces, or executable instructions to VLA / WAM executors. NovaPlan~\cite{fu2026novaplan} combines a closed-loop VLM planner with generated video references and geometrically grounded execution for zero-shot long-horizon manipulation. These works highlight the value of memory and subtask reasoning, but typically rely on VLA policies, external planners, or explicit intermediate plans. WorldScape Policy 2.0 instead integrates recent visual dynamics, event-level progress memory, and latent subgoal reasoning within a controllable WAM.

\subsection{Fine-Grained Prompts for Steerable Policies}

Fine-grained language prompts provide an important control interface for steerable manipulation, since coarse trajectory-level instructions cannot specify diverse atomic actions, object states, execution styles, and intermediate subgoals. At the generalist-model level, Qwen-VLA~\cite{Qwen-VLA} unifies continuous action and trajectory generation across tasks, environments, and robot embodiments through embodiment-aware prompting, while retaining fine-grained control within the same policy. FineVLA~\cite{Fine-VLA} instead isolates language granularity as a supervision axis, aligning goal-level and process-level instructions with robot trajectories to improve factor-level steerability. Related works introduce dense control or grounding signals via pixel-level visual prompts~\cite{PixelVLA}, 3D spatial action representations~\cite{SpatialVLA}, reconstruction-based visual grounding~\cite{ReconVLA}, and auxiliary spatial co-training~\cite{SG-VLA}.

Visual prompting provides a complementary interface for specifying where and how to act. TraceVLA~\cite{zheng2024tracevla} renders state-action trajectories as visual traces to improve spatiotemporal grounding, whereas CoT-VLA~\cite{zhao2025cotvla} autoregressively predicts future subgoal images as visual reasoning steps before action generation. Human-demonstration-conditioned policies~\cite{zhu2025humanvideoprompt} further use a single human video as a task prompt and align human and robot representations for cross-embodiment transfer. Explicit subtask decomposition can also provide fine-grained executable instructions to WAM executors~\cite{DSWAM}. WorldScape Policy 2.0 unifies these complementary directions through event-grounded WAM pretraining, aligning fine-grained subtask captions, goal images, and video demonstrations with visual state transitions and atomic action trajectories in ManipEvent-5M.

\section{Method}
\label{sec:method}

\begin{figure*}[t]
\centering
\includegraphics[width=16cm]{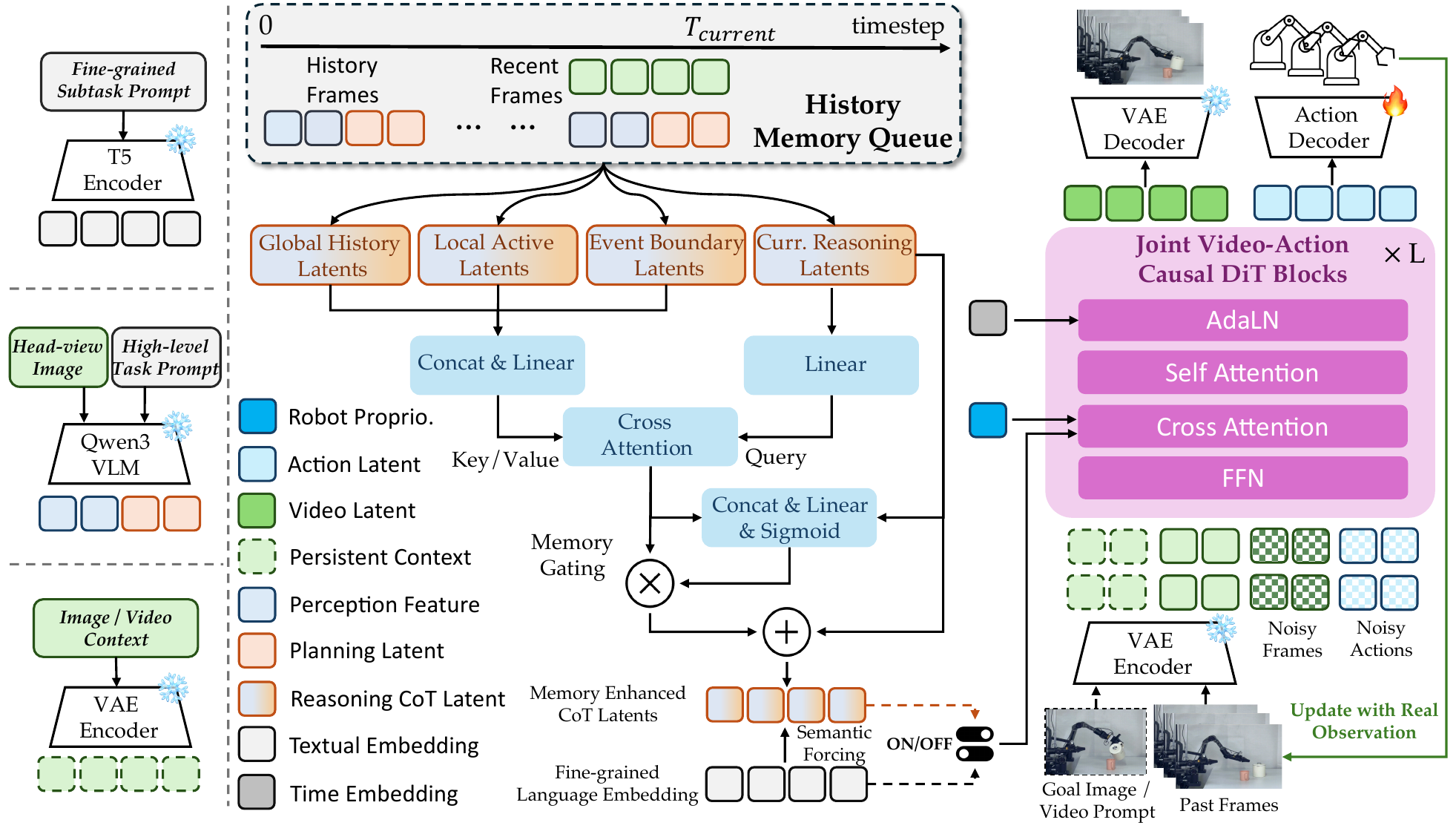}
\vspace{-0.2cm}
\caption{\textbf{Illustration of Reasoning-Augmented Long Short-Term Memory.} Historical VLM outputs form global-history, local-active, and event-boundary memories, which are retrieved and gated to augment the current reasoning tokens; semantic forcing aligns them with subtask semantics. Recent observations and optional visual prompts enter the DiT as causal visual prefill, while memory-enhanced VLM tokens or T5 embeddings condition joint video-action flow prediction according to the interaction mode.}
\label{fig:lstm}
\end{figure*}



\subsection{Overview and Problem Formulation}
\label{sec:problem_formulation}

WorldScape Policy 2.0 is a controllable World Action Model designed to support autonomous planning and interactive manipulation within a shared video-action backbone. Let $o_t$ denote the multi-view observation at time $t$, $a_t$ the corresponding robot action, and $z_t$ the VAE-encoded visual latent of $o_t$. We distinguish two language interaction modes, $\mu\in\{\mathrm{auto},\mathrm{fine}\}$, whose language inputs are an episode-level instruction $y$ and an event-level instruction $c_t$, respectively. The corresponding prompt set is $\mathcal{P}_t^{\mu}=\{p_{\mathrm{lang}}^{\mu},p_{\mathrm{goal}},p_{\mathrm{video}}\}$, where $p_{\mathrm{lang}}^{\mathrm{auto}}=y$ and $p_{\mathrm{lang}}^{\mathrm{fine}}=c_t$. Historical context is summarized by $\mathcal{M}_t=(\mathcal{Q}_t,\mathcal{Z}_t^{\mathrm{vis}})$, comprising the event-memory queue $\mathcal{Q}_t$ and the recent visual-latent buffer $\mathcal{Z}_t^{\mathrm{vis}}$. The event queue is activated for autonomous planning during the mid-training stage (see Sec.~\ref{sec:training_objective}) and is empty when event memory is disabled, whereas the visual buffer is used throughout all training stages. The model predicts an $H$-step action chunk and $H$ future visual latents as:
\begin{equation}
p_{\theta}(A_t^{(e)},z_{t+1:t+H}\mid o_t,a_{<t},\mathcal{P}_t^{\mu},\mathcal{M}_t).
\label{eq:wam_objective}
\end{equation}

\noindent\emph{(1) Unified action representation.}
For robot embodiment $e$, actions are expressed relative to the first frame of each action chunk. For a chunk beginning at time $t$ and future offset $\tau\in\{1,\ldots,H\}$, each per-arm action is
\begin{equation}
a_{t,\tau}^{(e)}
=
\left[\Delta p_{t,\tau}^{(e)};\phi_6\!\left(\Delta R_{t,\tau}^{(e)}\right);g_{t+\tau}^{(e)}\right]
\in\mathbb{R}^{10},
\quad
\Delta p_{t,\tau}^{(e)}=p_{t+\tau}^{(e)}-p_t^{(e)},
\quad
\Delta R_{t,\tau}^{(e)}=\left(R_t^{(e)}\right)^\top R_{t+\tau}^{(e)}.
\label{eq:relative_action}
\end{equation}
Here, $p_s^{(e)}\in\mathbb{R}^{3}$ and $R_s^{(e)}\in\mathrm{SO}(3)$ are the Cartesian position and orientation at time $s$, and $\phi_6:\mathrm{SO}(3)\rightarrow\mathbb{R}^{6}$ converts the relative rotation to its continuous 6D representation. Thus, the position and rotation encode a delta pose relative to the chunk's first frame, whereas $g_{t+\tau}^{(e)}\in\mathbb{R}$ is the absolute one-DoF gripper command. We collect the future controls as $A_t^{(e)}=(a_{t,\tau}^{(e)})_{\tau=1}^{H}\in\mathbb{R}^{H\times d_a}$, where $d_a=20$ for a dual-arm embodiment formed by concatenating two 10D per-arm vectors.

\noindent\emph{(2) Embodiment-specific action adapters.}
Each embodiment has a dedicated action encoder $E_a^{(e)}$ and action decoder $D_a^{(e)}$:
\begin{equation}
\widetilde{A}_{\rho}^{(e)}=E_a^{(e)}(A_{\rho}^{(e)}),\qquad
\hat v_{\theta}^{A,(e)}=D_a^{(e)}(h_{\theta}^{A,(e)}),
\label{eq:action_adapters}
\end{equation}
where $E_a^{(e)}$ projects a flow-perturbed raw action chunk into the shared DiT token space and $D_a^{(e)}$ maps the resulting action hidden states back to a velocity field in the raw action space. The causal video-action DiT is shared across embodiments, while these adapters absorb embodiment-specific kinematics, action scales, and control conventions. Importantly, the adapters only provide the input and output interfaces: flow matching and flow integration are both performed on the raw action chunk rather than on $\widetilde{A}_{\rho}^{(e)}$.

As shown in Fig.~\ref{fig:framework}, language and visual prompts follow distinct conditioning paths. Fine-grained instructions are encoded by T5, high-level instructions are interpreted jointly with the current observation by the VLM reasoning branch, and visual prompts are encoded as persistent VAE prefill. The resulting mode-specific conditions are integrated with event and visual memory before the DiT jointly predicts future video and action chunks.

\subsection{Event-Grounded World Action Modeling}

Existing WAMs are often trained with coarse episode-level instructions or a single interaction modality, providing weak supervision for the diverse state changes and atomic actions inside long manipulation episodes. WorldScape Policy 2.0 instead grounds WAM learning at the event level, where each segment aligns fine-grained subtask semantics with visual transitions and paired robot actions. This event-level structure supports two language interaction modes and complementary visual prompting. A fine-grained subtask instruction is encoded by T5 and directly controls the WAM, whereas a high-level task instruction is interpreted by the VLM branch to support autonomous subgoal planning. Goal images and video demonstrations provide additional persistent visual context. We formulate the encoded prompt as:
\begingroup
\small
\begin{equation}
s_t = E_{\mathrm{T5}}(c_t), \quad
z_t^{\mathrm{goal}} = E_{\mathrm{VAE}}(p_{\mathrm{goal}}), \quad
v_t = E_{\mathrm{VAE}}(p_{\mathrm{video}}),
\label{eq:prompt_tokens}
\end{equation}
\endgroup
where $c_t$ is a fine-grained subtask instruction, $s_t$ is its T5 embedding, and $z_t^{\mathrm{goal}}$ and $v_t$ are clean VAE latents of goal images and video demonstrations. The persistent visual condition is selected according to the available visual prompt:
\begin{equation}
p_t^{\mathrm{vis}}=
\begin{cases}
z_t^{\mathrm{goal}}, & \text{goal-image prompting},\\
v_t, & \text{video-demonstration prompting},\\
[z_t^{\mathrm{goal}};v_t], & \text{both prompts available},\\
\varnothing, & \text{otherwise}.
\end{cases}
\label{eq:visual_prompt}
\end{equation}
The language condition is likewise selected by $\mu$, rather than concatenating the T5 and VLM branches into a single input.

\noindent
\textbf{Mode-dependent language grounding.} In fine-grained instruction following, the segment-level caption $c_t$ explicitly specifies the atomic action or intermediate subgoal. Its T5 embedding $s_t$ is directly injected into the DiT through cross-attention, enabling precise and steerable action execution. In autonomous planning, the robot receives only an episode-level high-level instruction $y$. The VLM jointly interprets $y$, the current observation, and historical event memory to infer the active subgoal; the resulting memory-enhanced VLM tokens, rather than a T5 embedding of $y$, condition the DiT. Event-level captions supervise this latent planning pathway through semantic forcing during training.

\noindent
\textbf{Goal-conditioned policy learning.} Goal image prompts specify the desired target state visually. WorldScape Policy 2.0 encodes the goal image as $z_t^{\mathrm{goal}}$ and retains it in the causal visual prefill, allowing flow-perturbed future chunks to attend to the desired state throughout generation. The model is thus trained to infer an executable action path from the current observation to the visual goal.

\noindent
\textbf{Demonstration-conditioned in-context adaptation.} Video context prompts introduce cross-embodiment adaptation tasks, where the context may come from human-to-robot demonstrations, robot-to-robot demonstrations, or previous contextual episodes with different embodiments and viewpoints. WorldScape Policy 2.0 represents these demonstrations as $v_t$ and retains them as persistent visual prefill context alongside recent observations. This training mode encourages the model to infer object correspondences, transferable action patterns, and latent task rules from visual demonstrations, directly matching the video-context annotations provided by ManipEvent-5M dataset introduced in Sec.~\ref{sec:manipevent_dataset}.

\subsection{Reasoning-Augmented Long Short-Term Memory}

Long-horizon manipulation requires memory at two complementary representation levels. At the event level, the model must retain semantic evidence about completed subtasks and retrieve history relevant to the current task stage. At the visual level, it must preserve recent frame-level dynamics, such as contact changes and object motion, for temporally coherent video-action prediction. As illustrated in Fig.~\ref{fig:lstm}, WorldScape Policy 2.0 therefore combines a \emph{long short-term event memory} in the VLM reasoning branch with a \emph{short-term visual memory} in the causal DiT branch. The former performs progress-aware retrieval over historical chunk representations, whereas the latter directly supplies recent visual latents as causal prefill context. Optional goal images and video demonstrations are further retained as persistent visual context throughout the rollout.

\noindent
\textbf{Reasoning latent formulation.}
For each action chunk under autonomous planning, perception and planning share a single VLM context rather than using separate forward passes. The context combines the current head-view observation $o_t^{\mathrm{head}}$ with a task-conditioned planning prompt $\widetilde{y}_t=T_{\mathrm{plan}}(y)$. We instantiate $T_{\mathrm{plan}}$ with the following fixed template, where \texttt{\{task\}} is replaced by the episode-level instruction $y$:
\begin{quote}
\small\itshape
``You are a robot planner. Instructions: \{task\}. Given the current high-level task instruction and current head-view observation, predict the next atomic action subtask for the next second.''
\end{quote}
A single VLM prefill jointly encodes $(o_t^{\mathrm{head}},\widetilde{y}_t)$ and produces the final-layer context hidden states $H_t^{\mathrm{ctx}}$ together with their KV cache. The perception tokens are read directly from this VLM output:
\begin{equation}
u_t = H_t^{\mathrm{ctx}}
=H_{\mathrm{VLM}}^{L}(o_t^{\mathrm{head}},\widetilde{y}_t).
\label{eq:perception_tokens}
\end{equation}
Starting from the same cached context, the VLM then continues autoregressively and generates $K$ discrete planning tokens using greedy decoding:
\begin{equation}
w_t^k=\arg\max_{w}\,p_{\mathrm{VLM}}
(w\mid o_t^{\mathrm{head}},\widetilde{y}_t,w_t^{<k}),
\quad k=1,\ldots,K.
\label{eq:planning_tokens}
\end{equation}
The final-layer hidden state returned online at each decoding step is retained as the corresponding latent planning token:
\begin{equation}
r_t^k=\left[
H_{\mathrm{VLM}}^{L}
(o_t^{\mathrm{head}},\widetilde{y}_t,w_t^{1:k})
\right]_{\operatorname{pos}(w_t^k)},
\quad k=1,\ldots,K.
\label{eq:planning_features}
\end{equation}
Thus, $u_t$ and $r_t^{1:K}$ belong to one continuous VLM inference trajectory: $u_t$ encodes the shared visual-language prefill, while $r_t^{1:K}$ summarizes its autoregressively inferred task continuation. We concatenate the two token groups to form the current reasoning latent
\begin{equation}
q_t = [u_t; r_t^{1:K}],
\label{eq:reasoning_latent}
\end{equation}
which jointly represents the current state and the inferred next subgoal. Historical observations are processed using the same perception and planning branches, ensuring that every memory entry has the same two-part token structure.

\noindent
\textbf{Long short-term event memory construction.}
The event-memory branch maintains the queue $\mathcal{Q}_t=\{q_1,\ldots,q_{t-1}\}$ of historical chunk-level VLM outputs introduced above. Each $q_j$ contains the perception tokens and latent planning tokens associated with chunk $j$. To control the cost of full-history retrieval, we compress the perception tokens of every historical chunk into a fixed number of learned gist tokens through attention pooling, while retaining its planning tokens. Concatenating the resulting tokens gives a compact full-history bank $H_t$. We then explicitly construct three complementary memory views. \textbf{Global-History latents} summarize the complete trajectory, \textbf{Local-Active latents} represent the currently active event, and \textbf{Event-Boundary latents} sparsely preserve major semantic transitions:
\begin{equation}
\begin{array}{l}
m_t^{\mathrm{gh}}=P_{\mathrm{gh}}\!\left(\operatorname{Expand}
\left(F_{\mathrm{gh}}([e_y;\operatorname{Mean}(H_t)])\right)\right),\\
m_t^{\mathrm{la}}=P_{\mathrm{la}}([q_j\mid j\in\mathcal{I}_t^{\mathrm{la}}]),\quad
\mathcal{I}_t^{\mathrm{la}}=\{\max(1,t-S_e),\ldots,t-1\},\\
m_t^{\mathrm{eb}}=P_{\mathrm{eb}}([q_j\mid j\in\mathcal{B}_t]),
\end{array}
\label{eq:memory_candidates}
\end{equation}
Here, $e_y$ is the high-level instruction embedding, $\operatorname{Mean}$ averages over the token dimension of the compact history bank $H_t$, $F_{\mathrm{gh}}$ fuses task intent with pooled history, and $\operatorname{Expand}$ produces a fixed number of global-history slots. The learned projections $P_{\mathrm{gh}}$, $P_{\mathrm{la}}$, and $P_{\mathrm{eb}}$ map the three memory views into a shared feature space. The notation $[q_j\mid j\in\mathcal{S}]$ denotes temporally ordered token concatenation over index set $\mathcal{S}$; $\mathcal{I}_t^{\mathrm{la}}$ contains the latest $S_e$ completed chunks, where $S_e$ is the local-active window length, and $\mathcal{B}_t$ is the event-boundary index set defined below. The recent local-active and selected boundary chunks are retained as full-token anchors, preserving fine-grained perception and planning evidence.

Event-boundary chunks are selected directly from latent changes in the historical VLM outputs. Let $\bar q_j=\operatorname{Mean}(q_j)$ denote the token-averaged representation of chunk $j$. We compute
\begin{equation}
d_j=1-\operatorname{cos}(\bar q_j,\bar q_{j-1}),\qquad
\mathcal{B}_t=\operatorname{TopK}_{\Delta}(\{d_j\}_{j=2}^{t-1};S_b),
\label{eq:event_boundary}
\end{equation}
where $d_j$ measures the semantic change between consecutive chunks. $\operatorname{TopK}_{\Delta}$ greedily selects at most $S_b$ highest-change indices while enforcing a minimum temporal separation $\Delta$ between selected indices. The resulting event-boundary latents sparsely retain states at which a subtask likely starts, terminates, or changes, without requiring explicit online boundary annotations. As with local-active memory, the selected chunks are stored as full-token anchor frames, allowing the model to revisit detailed perception and planning evidence at these critical transitions rather than only their pooled boundary scores. Consequently, event memory combines long-term task-progress context from global-history and event-boundary latents with short-term semantic context from local-active latents. The anchor-frame implementation preserves local detail within this semantic memory organization, which remains distinct from the frame-level visual memory described below.

\noindent
\textbf{Event-memory retrieval and gated fusion.}
Following the implementation, the three specialized memory views are concatenated with the compact full-history bank, ensuring that selection does not discard non-boundary evidence. Let
\begin{equation}
B_t=\operatorname{cat}_{\mathrm{tok}}
\left(m_t^{\mathrm{gh}},m_t^{\mathrm{la}},m_t^{\mathrm{eb}},H_t\right)
\end{equation}
denote the concatenated memory tokens. For $q_t\in\mathbb{R}^{N_q\times d_{\mathrm{VLM}}}$ and $B_t\in\mathbb{R}^{N_m\times d_{\mathrm{VLM}}}$, we form independent query, key, and value projections and retrieve memory by
\begin{equation}
\begin{aligned}
Q_t&=P_q(q_t),\qquad K_t=P_k(B_t),\qquad V_t=P_v(B_t),\\
\mathcal{A}_t^{\mathrm{mem}}
&=\operatorname{softmax}_{\mathrm{mem}}\!\left(
\frac{Q_tK_t^{\top}}{\sqrt{d_k}}+M_t\right),\\
\widetilde{m}_t&=P_o\!\left(\mathcal{A}_t^{\mathrm{mem}}V_t\right)
\in\mathbb{R}^{N_q\times d_{\mathrm{VLM}}},
\end{aligned}
\label{eq:memory_weight}
\end{equation}
where $Q_t\in\mathbb{R}^{N_q\times d_k}$, $K_t\in\mathbb{R}^{N_m\times d_k}$, $V_t\in\mathbb{R}^{N_m\times d_v}$, $M_t\in\mathbb{R}^{N_q\times N_m}$ masks padded history and boundary slots, and $P_o:\mathbb{R}^{d_v}\rightarrow\mathbb{R}^{d_{\mathrm{VLM}}}$ maps the attended values back to the VLM hidden dimension.
Because historical evidence is not equally useful at every step, a learned gate controls how much retrieved memory is injected into each current token:
\begin{equation}
\gamma_t = \sigma\!\left(W_g[q_t;\widetilde{m}_t]+b_g\right),
\qquad
\hat{q}_t = q_t + \alpha\,\gamma_t \odot \widetilde{m}_t,
\label{eq:memory_fusion}
\end{equation}
where $\sigma$ is the sigmoid function, $W_g:\mathbb{R}^{2d_{\mathrm{VLM}}}\rightarrow\mathbb{R}$ and $b_g$ are learned gate parameters applied to each token, $[\cdot;\cdot]$ denotes feature-wise concatenation, and $\odot$ denotes element-wise multiplication with broadcasting over the feature dimension. Thus, $\gamma_t\in\mathbb{R}^{N_q\times1}$ is a token-wise memory-importance gate, $\alpha$ is a residual scaling factor, and $\hat{q}_t$ is the memory-enhanced reasoning latent. This retrieval mechanism allows the model to selectively use task-level progress, recent event context, sparse transition evidence, and the compressed full history.

\begin{wrapfigure}[21]{r}{0.52\textwidth}
  \centering
  \includegraphics[width=\linewidth]{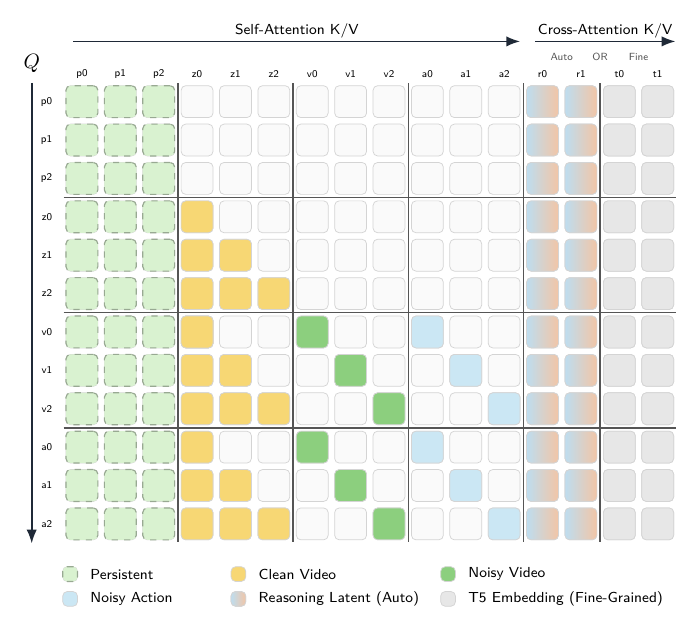}
  \caption{\textbf{Attention mask of WorldScape Policy 2.0.}}
  \label{fig:attention_mask_worldscape}
\end{wrapfigure}

\noindent
\textbf{Visual short-term memory and persistent context.}
In parallel to event-memory retrieval, the causal DiT models recent frame-level dynamics through visual prefilling. We define the recent visual-latent buffer as $\mathcal{Z}_t^{\mathrm{vis}}=z_{t-S_v:t}^{\mathrm{obs}}$, containing the clean VAE latents of the current observation and the most recent $S_v$ visual chunks. These latents precede the flow-interpolated future video latents and embodiment-encoded action tokens in the DiT sequence:
\begin{equation}
h_t^{\mathrm{DiT},(e)}=
[p_t^{\mathrm{vis}};\mathcal{Z}_t^{\mathrm{vis}};z_{t+1:t+H}^{\rho};
E_a^{(e)}(A_{\rho,t}^{(e)})],
\end{equation}
where $\mathcal{Z}_t^{\mathrm{vis}}$ is updated online after each executed action chunk and $E_a^{(e)}$ is selected according to the current embodiment. As summarized in Fig.~\ref{fig:attention_mask_worldscape}, a causal visual-attention mask allows future video-action tokens to attend to persistent visual prompts and available clean history while preventing information leakage from future chunks. This pathway supplies local motion and interaction continuity through DiT self-attention, rather than first compressing recent frames into the VLM event memory.

The optional prompt latents $p_t^{\mathrm{vis}}$ remain fixed across rollout steps, unlike the sliding recent-frame buffer, allowing the model to continuously reference the desired goal state or demonstrated trajectory. Both recent and persistent visual latents enter the DiT through causal self-attention, while the cross-attention condition is selected by interaction mode: T5 embeddings are used for fine-grained instruction following, whereas memory-enhanced VLM tokens are used for autonomous planning. This separation preserves the respective roles of semantic event retrieval, local visual dynamics, and multimodal task conditioning.

\begin{figure*}[t]
  \centering
  \includegraphics[width=16.2cm]{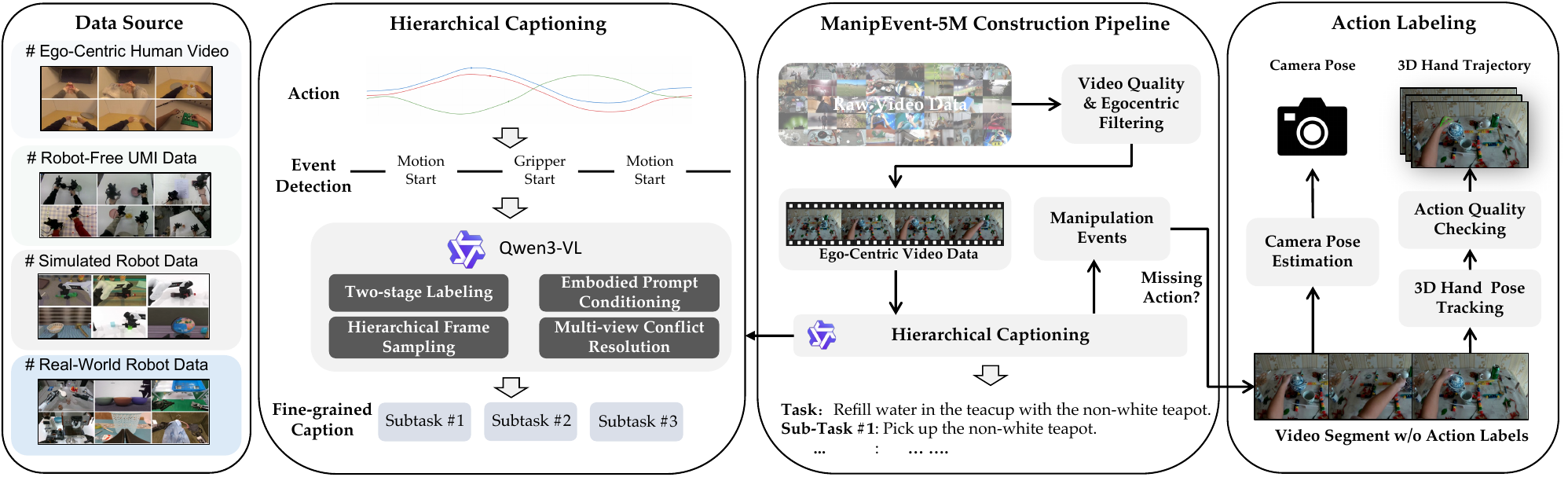}
  \caption{\textbf{Construction pipeline of ManipEvent-5M}. We aggregate heterogeneous manipulation sources, including human-arm ego videos, robot-free UMI data, simulated trajectories, and real-robot demonstrations, and convert them into event-level training samples. Each trajectory is segmented into ordered atomic subgoals and annotated with paired action trajectories, episode-level global instructions, segment-level fine-grained captions, goal-image prompts, and video-context prompts. This event-based multimodal schema supports the event-grounded WAM interface in Fig.~\ref{fig:framework}, enabling fine-grained language grounding, goal-conditioned policy learning, demonstration-conditioned in-context learning, and long-horizon memory-aware mid-training.}
  \label{fig:pipeline}
  \end{figure*}

\subsection{Implicit Subgoal Latent Planning}

WorldScape Policy 2.0 uses event-level subtask captions in two different ways. In fine-grained instruction following, the T5 embedding $s_t=E_{\mathrm{T5}}(c_t)$ acts as the direct language condition. In autonomous planning, the caption is used only as training-time semantic supervision for the VLM planning branch, which produces $\hat{q}_t$ from perception, autoregressive planning tokens, and retrieved historical memory.

\noindent
\textbf{Subgoal semantic forcing.}
During training, $s_t$ provides a semantic target for the autonomous-planning latent. Because $s_t$ and $\hat{q}_t$ are token sequences of potentially different lengths, we first obtain fixed-dimensional summaries $\bar{s}_t=\operatorname{Pool}(s_t)$ and $\bar{q}_t=\operatorname{Pool}(\hat{q}_t)$. We keep the T5 target encoder fixed and apply stop-gradient to its normalized representation, while a trainable projector $\phi_q$ maps the reasoning summary into the T5 semantic space:
\begin{equation}
\begin{aligned}
\widetilde{s}_t&=\operatorname{sg}\!\left(
\frac{\bar{s}_t}{\max(\|\bar{s}_t\|_2,\varepsilon)}\right),\\
\widetilde{q}_t&=
\frac{\phi_q(\bar{q}_t)}
{\max(\|\phi_q(\bar{q}_t)\|_2,\varepsilon)},\\
\mathcal{L}_{\mathrm{sem}}&=1-\widetilde{s}_t^{\top}\widetilde{q}_t,
\end{aligned}
\label{eq:semantic_forcing}
\end{equation}
where $\operatorname{sg}$ denotes stop-gradient and $\varepsilon>0$ prevents division by zero. Fixing the semantic target prevents the alignment objective from drifting toward a jointly collapsed solution, while $\phi_q$ learns to align the reasoning latent with explicit subtask semantics. During autonomous inference, the fine-grained caption and its T5 embedding are absent; the model receives only the high-level instruction and relies on $\hat{q}_t$ to infer the active subgoal from the current observation and event memory.

\noindent
\textbf{Mode-dependent video-action prediction.}
The language-side cross-attention condition is selected according to the requested interaction mode:
\begin{equation}
\ell_t=
\begin{cases}
s_t, & \text{fine-grained instruction following},\\
\hat{q}_t, & \text{autonomous planning},
\end{cases}
\end{equation}
where $s_t$ provides explicit atomic-action control, while $\hat{q}_t$ provides an implicitly inferred and progress-aware subgoal condition. Let $\rho$ denote the flow timestep embedded through AdaLN. The shared causal DiT produces action and video hidden states, after which the embodiment-specific action decoder and shared video output head predict their respective velocity fields:
\begin{equation}
\begin{aligned}
(h_{\theta}^{A,(e)},h_{\theta}^{z})
&=F_{\theta}(h_t^{\mathrm{DiT},(e)},\ell_t,\rho),\\
\hat v_{\theta}^{A,(e)}&=D_a^{(e)}(h_{\theta}^{A,(e)}),\qquad
\hat v_{\theta}^{z}=D_z(h_{\theta}^{z}),
\end{aligned}
\label{eq:decoder}
\end{equation}
where $F_{\theta}$ denotes the shared causal video-action DiT, $D_z$ is the video output head, and visual prompts, when available, are already included in $h_t^{\mathrm{DiT},(e)}$ as persistent prefill context. At inference, the video field is integrated in VAE-latent space, whereas $\hat v_{\theta}^{A,(e)}$ is integrated directly in the raw action space from $\rho=1$ (noise) to $\rho=0$ (data). At each flow step, the current raw action chunk is re-encoded by $E_a^{(e)}$ before the next velocity evaluation. In autonomous planning, the model repeatedly updates event and visual memory, infers $\hat{q}_t$, predicts the next chunk, and rolls forward with new observations. In fine-grained instruction following, the externally specified $s_t$ directly steers each atomic action without requiring the model to infer the subgoal from the high-level instruction alone.

\subsection{ManipEvent-5M: Event-Based Multimodal Dataset}
\label{sec:manipevent_dataset}
To facilitate native pretraining of WorldScape Policy 2.0, we construct \textbf{ManipEvent-5M}, the 
event-based multimodal dataset illustrated in Fig.~\ref{fig:pipeline}. It aggregates human-arm 
egocentric videos, robot-free UMI demonstrations, simulated trajectories, and real-robot data from 
self-collected and public sources. Each trajectory is decomposed into an ordered sequence of events:
\begin{equation}
\tau=\{(o_{t_b^j:t_e^j},a_{t_b^j:t_e^j},y,c_j,
p_{\mathrm{goal}}^j,p_{\mathrm{video}}^j)\}_{j=1}^{N_{\tau}},
\label{eq:event_dataset}
\end{equation}
where $(t_b^j,t_e^j)$ is the temporal boundary of event $j$, $y$ is the episode-level instruction, $c_j$ is its fine-grained event caption, and $p_{\mathrm{goal}}^j$ and $p_{\mathrm{video}}^j$ are optional visual prompts. This schema separates global task intent from event-level subgoal semantics while preserving their alignment with observations and actions.

\noindent
\textbf{Action canonicalization and human-hand retargeting.}
Robot trajectories are converted to the Cartesian end-effector representation defined in Sec.~\ref{sec:problem_formulation}. For human data, estimated hand poses are retargeted to robot gripper poses and commands before pretraining, yielding action-aligned human demonstrations under the same position--rotation--gripper semantics. This canonical layout enables joint pretraining across human and robot data, while the embodiment-specific action encoders and decoders preserve distinct kinematic mappings and control conventions.

\begin{table*}[t]\footnotesize
\centering
\caption{\textbf{Statistics of WorldScape Policy 2.0 pretraining data in ManipEvent-5M.} The number of segments is reported for datasets with event-level decomposition. ``Avg. Seg. / Episode" is the number of event segments divided by the number of episodes, and ``Single-Seg. Ratio" is the proportion of episodes without further event decomposition.}
\label{tab:manipevent_stats}
\vspace{-0.2cm}
\resizebox{\textwidth}{!}{
\begin{tabular}{llccccccc}
\toprule[1pt]
\textbf{Data Type} & \textbf{Dataset Source} & \makecell{\textbf{Frames}\\\textbf{(M)}} & \makecell{\textbf{Duration}\\\textbf{(H)}} & \makecell{\textbf{Episodes}\\\textbf{(K)}} & \makecell{\textbf{Segments}\\\textbf{(M)}} & \makecell{\textbf{Avg. Seg.}\\\textbf{/ Episode}} & \makecell{\textbf{Single-Seg.}\\\textbf{Ratio (\%)}} & \makecell{\textbf{Training}\\\textbf{Ratio (\%)}} \\
\midrule
\makecell[l]{Real-World\\Robot Data} & Self-Collected PiPER Dataset & 54.00 & 506.60 & 35.82 & 0.56 & 15.6 & 0.5 & 45.0 \\
\midrule
\multirow{4}{*}{\makecell[l]{Public\\Robot Data}} & AgiBot World~\cite{bu2025agibot} & 285.56 & 2644.08 & 160.15 & 1.16 & 7.2 & 0.2 & 32.53 \\
 & RoboMIND~\cite{wu2025robomind} & 6.56 & 60.75 & 10.27 & -- & -- & 100.0 & 0.75 \\
 & RoboCOIN~\cite{wu2025robocoin} & 32.00 & 302.94 & 44.27 & -- & -- & 100.0 & 3.65 \\
 & DROID~\cite{Khazatsky-RSS-24} & 27.00 & 500.82 & 92.23 & -- & -- & 100.0 & 3.07 \\
\midrule
\makecell[l]{Robot-Free\\UMI Data} & Self-Collected UMI Dataset & 9.50 & 90.60 & 26.02 & 0.06 & 2.3 & 30.8 & 8.0 \\
\midrule
\multirow{2}{*}{\makecell[l]{Simulated\\Robot Data}} & RoboTwin 2.0~\cite{chen2025robotwin} & 8.16 & 42.16 & 35.73 & 0.145 & 4.1 & 26.8 & 3.94 \\
 & LIBERO~\cite{liu2023libero} & 0.12 & 3.86 & 1.71 & 0.005 & 2.9 & 0.12 & 0.06 \\
\midrule
\makecell[l]{Ego-Centric\\Human Data} & EgoDex~\cite{hoque2026egodex} & 89.24 & 831.00 & 338.23 & 2.96 & 8.8 & 18.8 & 3.0 \\
\midrule
\textbf{Total} &  & \textbf{512.14} & \textbf{4982.81} & \textbf{744.43} & \textbf{4.89} & \textbf{6.6} & -- & -- \\
\bottomrule[1pt]
\end{tabular}}
\end{table*}

\begin{figure*}[t]
\centering
\includegraphics[width=0.95\textwidth]{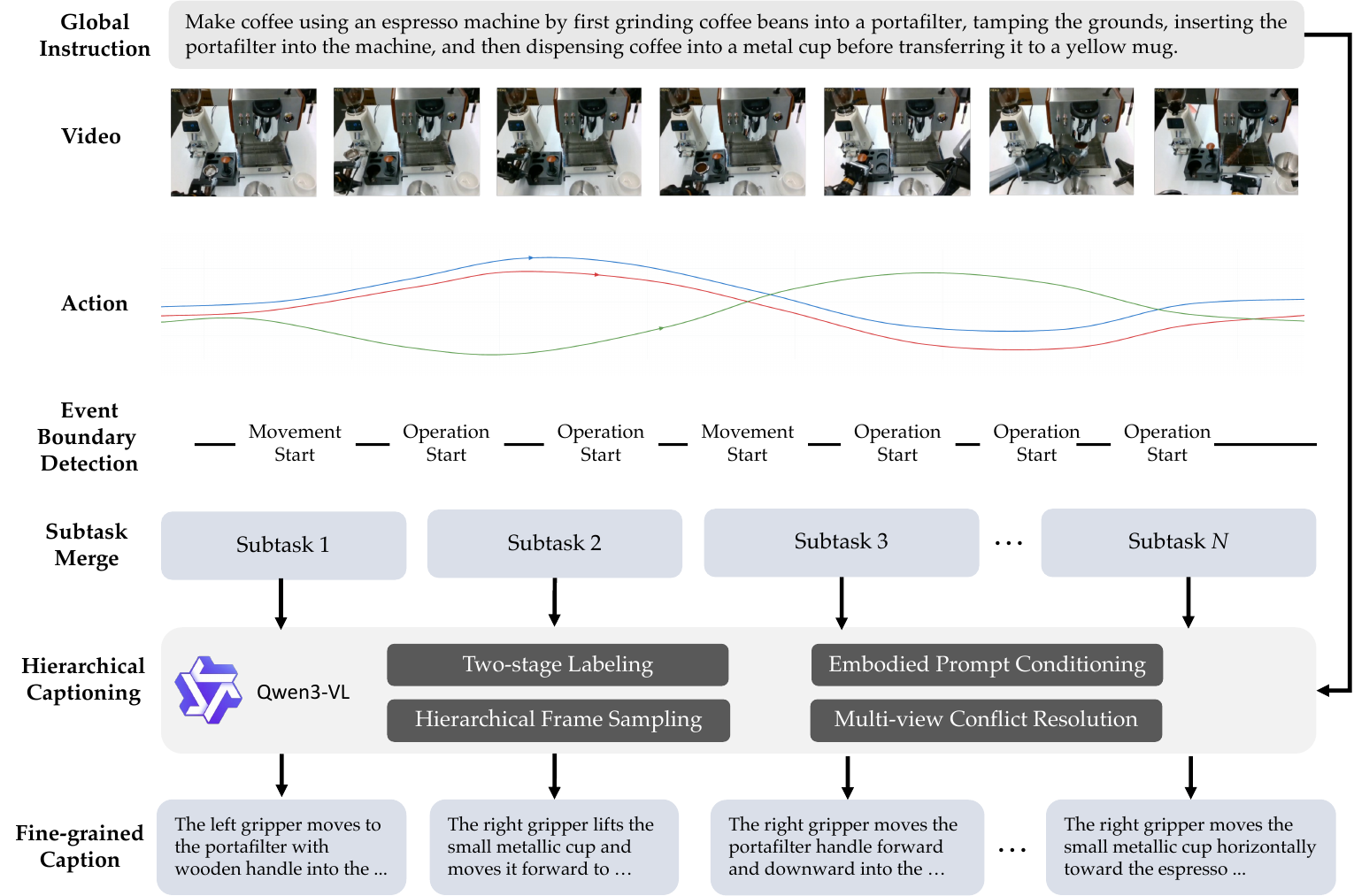}
\caption{\textbf{Examples of hierarchical captioning in ManipEvent-5M.} The pipeline combines two-stage temporal-semantic labeling, hierarchical episode / event frame sampling, embodied prompt conditioning, and multi-view conflict resolution. Representative instances map segmented events to fine-grained captions, complementing the construction pipeline in Fig.~\ref{fig:pipeline}.}
\label{fig:annotation_sample}
\end{figure*}

\noindent
\textbf{Text-prompt construction.}
We convert each trajectory into an episode-level task description and temporally aligned subtask captions using a four-stage hierarchical annotation procedure with Qwen3-VL~\cite{Qwen3-VL}, as illustrated in Fig.~\ref{fig:annotation_sample}.

\emph{(1) Two-stage labeling.}
We first establish temporal structure from robot-state signals and then perform open-vocabulary semantic relabeling with the VLM. For temporal segmentation, embodiment-normalized translational and rotational velocities are computed from end-effector or hand trajectories and smoothed. Motion-state transitions, gripper-state changes, and episode endpoints form candidate boundaries. Short segments are merged with adjacent events, and an adaptive granularity constraint prevents both semantically overloaded segments and fragments caused by sensor jitter. Each resulting event is assigned a coarse primitive prior, such as move, grasp / release, or idle, based on motion and gripper signals. Given these boundaries, the VLM determines the semantic content of the trajectory and each event. This two-stage design decouples \emph{when} an action changes from \emph{what} the action means, reducing missed steps, ordering errors, and temporal misalignment that arise when directly captioning a long video.

\emph{(2) Hierarchical frame sampling.}
Captioning is performed at both episode and event levels. At the episode level, we sample up to 16 representative timestamps from the full trajectory and provide their images together with the ordered event list, temporal boundaries, primitive-action priors, and execution outcome. Qwen3-VL returns a structured \texttt{high\_level} description that summarizes the task actually completed, i.e., a hindsight task description rather than a verbatim copy of the original command. This makes the label robust to missing instructions, execution deviations, and recovery attempts. At the event level, we uniformly sample up to 12 timestamps within each segment and collect their available camera views. Conditioned on the episode-level \texttt{high\_level}, event boundaries, primitive prior, and local visual evidence, the VLM generates a single-action \texttt{step\_text}. Separating global summarization from local captioning binds every instruction to a specific event while avoiding omissions and ordering errors in long action sequences.

\emph{(3) Embodied prompt conditioning.}
We translate the annotation guidelines into structured prompt constraints rather than requesting an unconstrained video summary. The prompt supplies the temporal metadata and primitive-action prior, and encourages each caption to identify the action order, acting end effector, target object, initial and terminal states, contact mode, motion direction, object interaction, body motion, and, when present, failure or recovery behavior. These constraints increase semantic density while retaining a natural-language action sentence; for example, the caption specifies which gripper interacts with which object and toward which target, rather than producing a generic description such as ``move the cup.''

\emph{(4) Multi-view conflict resolution.}
For event-level annotation, multi-view observations are presented with a local-first organization, \texttt{LEFT\_GRIPPER $\rightarrow$ RIGHT\_GRIPPER $\rightarrow$ HEAD}. The head camera provides global scene and task context, whereas wrist / gripper cameras offer more reliable evidence about contact, grasp state, and local target identity. The prompt therefore instructs the VLM to prioritize gripper-view evidence when it conflicts with the head view, mitigating target-object confusion in cluttered scenes. Finally, outputs are parsed and validated as structured records; invalid episode captions trigger a stricter retry or rule-based fallback, while a failed event caption is replaced locally without discarding valid labels from the remaining trajectory.

\begin{figure*}[t]
  \centering
  \includegraphics[width=0.95\textwidth]{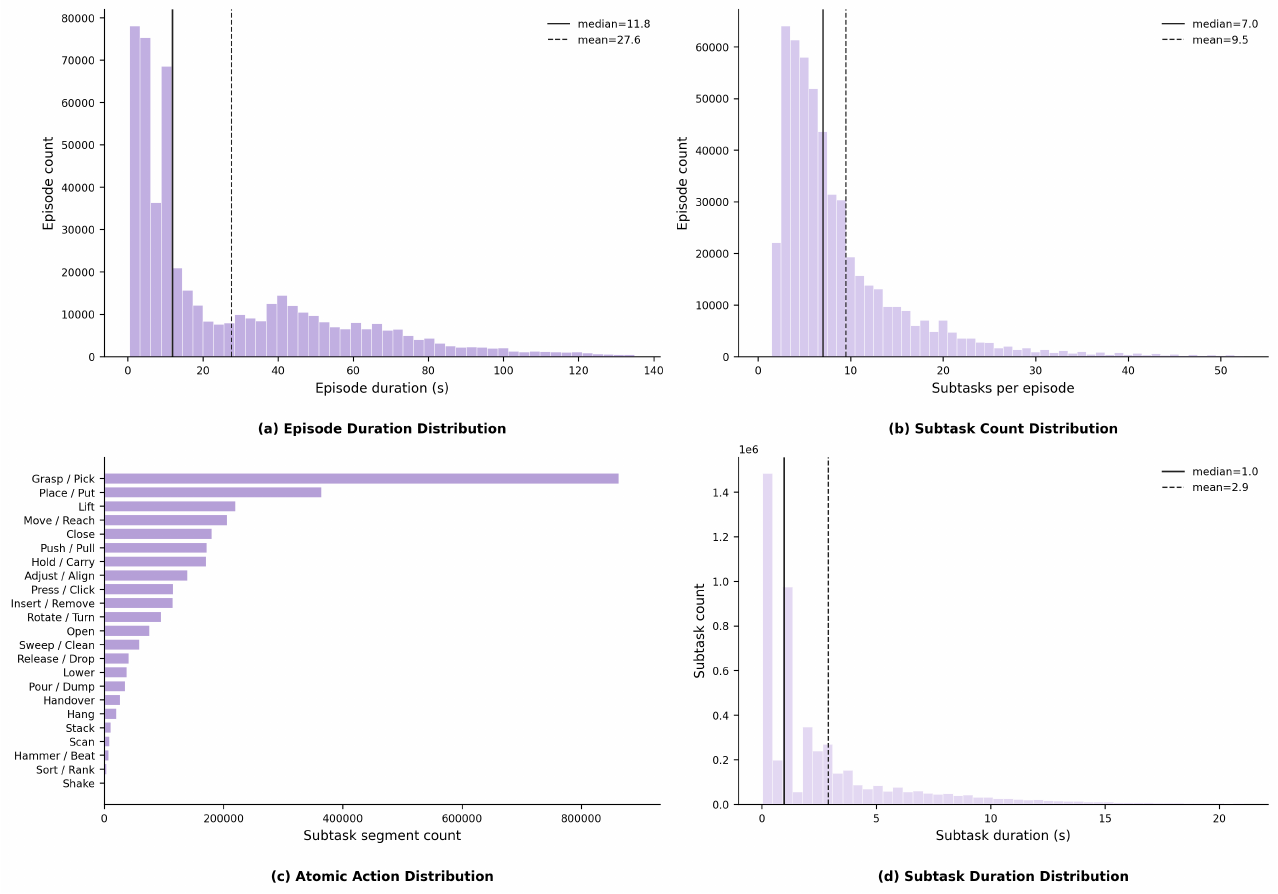}
  \caption{\textbf{Distribution statistics of ManipEvent-5M.} The figure shows the detailed distribution of episode duration, subtask count, atomic actions, and subtask duration across diverse sources from real-world / simulated robot data, robot-free data, and ego-centric human data in the constructed ManipEvent-5M dataset.}
  \label{fig:distribution}
  \end{figure*}

As summarized in Table~\ref{tab:manipevent_stats} and Fig.~\ref{fig:distribution}, ManipEvent-5M combines large-scale visual coverage with event-level subtask decomposition across heterogeneous data sources. Beyond fine-grained text annotations, we construct goal-image and video-demonstration prompts to support visual goal specification and cross-embodiment in-context adaptation.

\noindent
\textbf{Goal-image prompt construction.}
We construct two complementary types of goal-image prompts. A \emph{first-view goal image} is directly extracted from the terminal observation of each segmented event, depicting the desired post-condition from the same camera configuration and embodiment as the corresponding action trajectory. It therefore provides an unambiguous visual target for learning how the current state should transition after an atomic action. A \emph{third-view goal image} is sampled from a task-matched human demonstration at the completion of the corresponding event. Such images retain the human demonstrator and show the target object configuration from an external viewpoint. They require the policy to abstract away the demonstrator and viewpoint, infer the intended outcome, and reproduce the demonstrated state using the robot embodiment.

\noindent
\textbf{Video-prompt construction.}
Video prompts are designed primarily for human-to-robot and robot-to-robot task transfer. For human-to-robot transfer, we pair self-collected real-robot trajectories with UMI demonstrations collected for the same tasks. Pairs are matched by the global task, ordered event sequence, and terminal outcome, ensuring that the human and robot demonstrations express consistent action procedures and reducing ambiguity about which skill should be transferred. For robot-to-robot transfer, we collect or retrieve trajectories of the same semantic task executed by different robot embodiments and align them using their ordered event descriptions and goal states. The resulting context-target pairs encourage the model to identify embodiment-invariant object correspondences, action structure, and task intent while adapting execution to the target robot.

\begin{figure*}[t]
  \centering
  \includegraphics[width=0.95\textwidth]{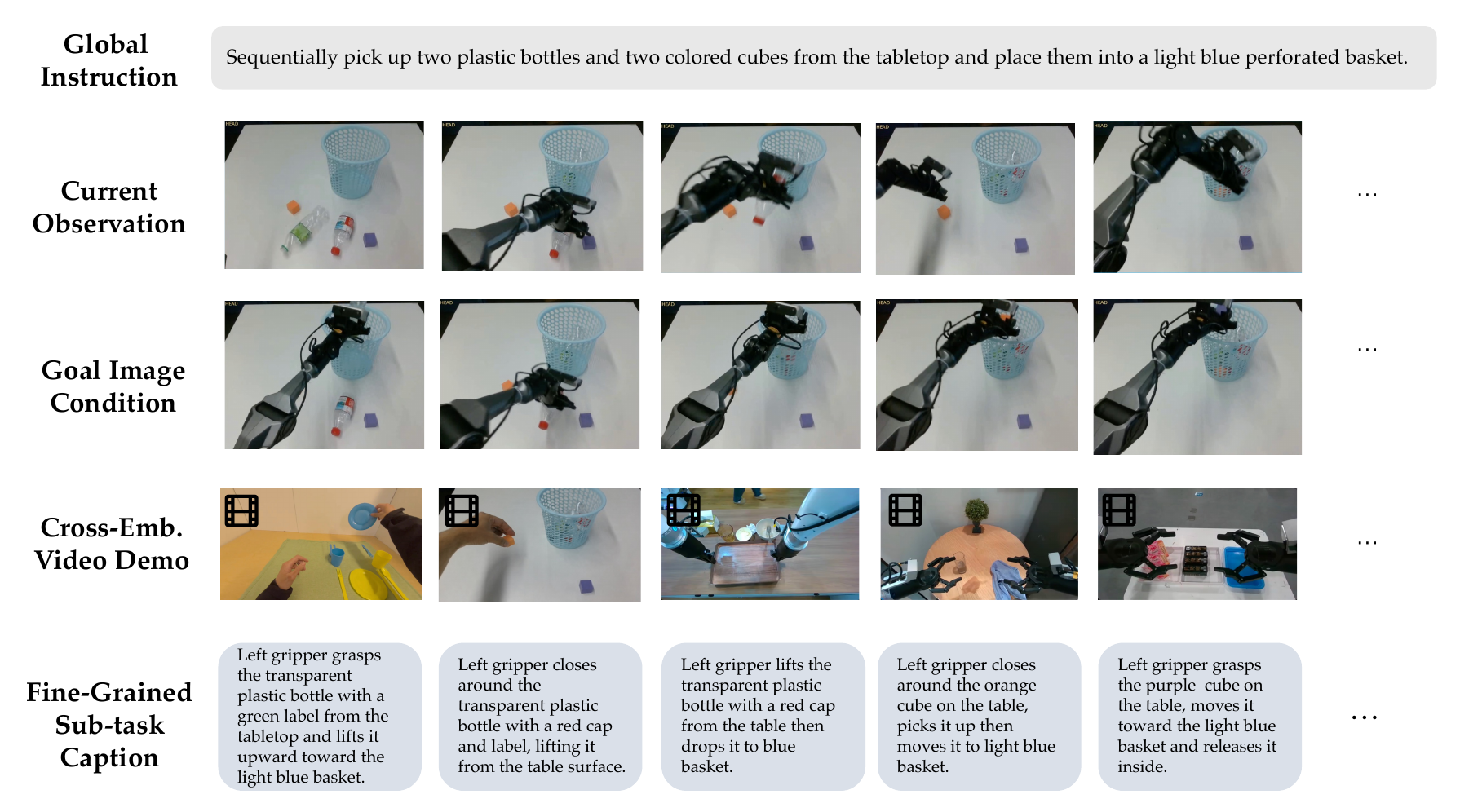}
  \caption{\textbf{Multimodal prompt example from ManipEvent-5M:} \texttt{desktop cleaning.} The sample pairs a global task instruction and current observations with goal-image conditioning, event-level fine-grained subtask captions, and a cross-embodiment video demonstration for placing bottles and colored cubes into a basket.}
  \label{fig:prompt_example_collection}
\end{figure*}

\begin{figure*}[t]
  \centering
  \includegraphics[width=0.95\textwidth]{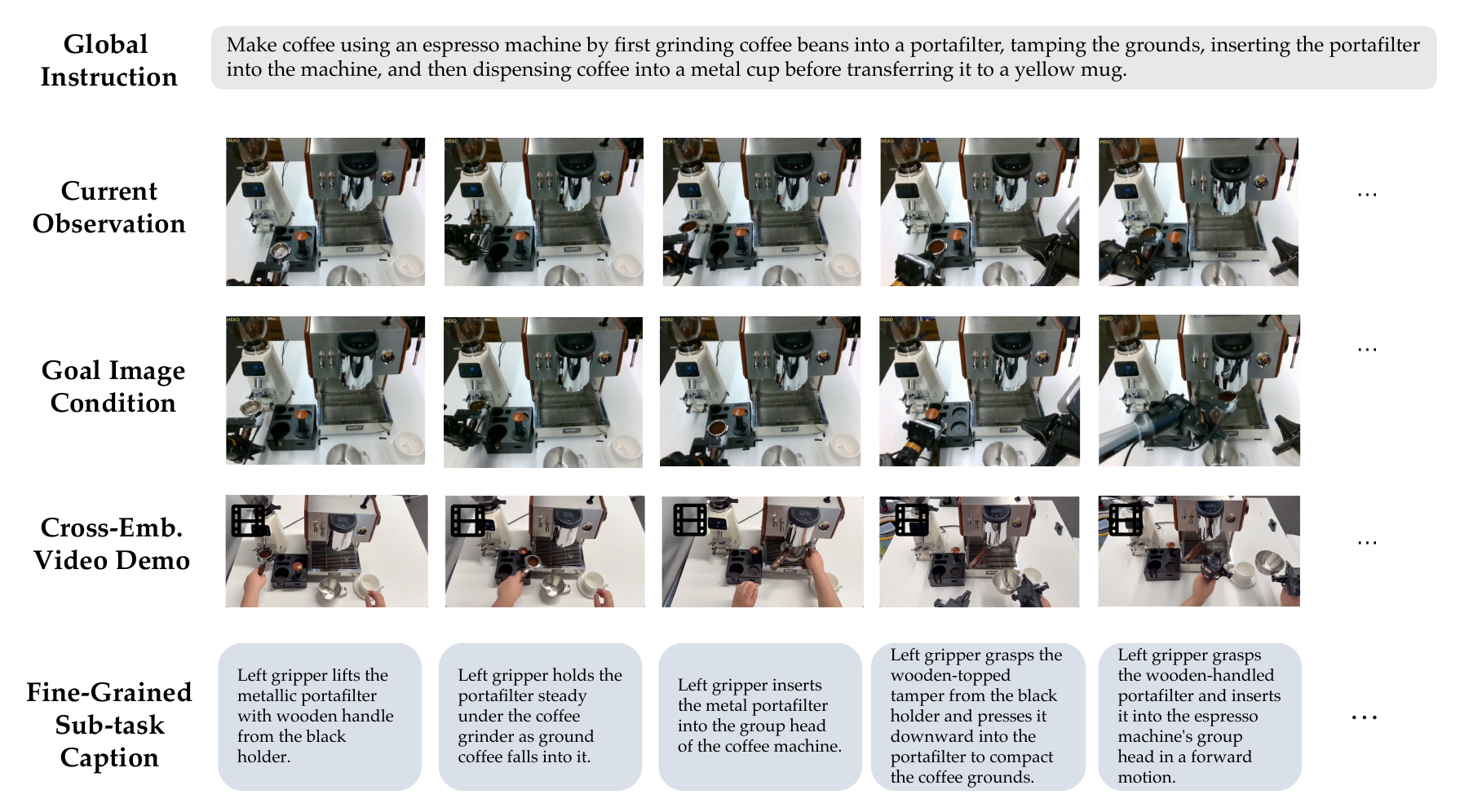}
  \caption{\textbf{Multimodal prompt example from ManipEvent-5M:} \texttt{coffee making.} The sample illustrates how a multi-stage task is represented by a global instruction, current observations, a goal-image condition, fine-grained subtask captions, and a cross-embodiment video demonstration.}
  \label{fig:prompt_example_coffee}
\end{figure*}

Figs.~\ref{fig:prompt_example_collection} and~\ref{fig:prompt_example_coffee} show representative multimodal prompt samples for \texttt{desktop cleaning} and long-horizon \texttt{coffee making}, respectively. Together, this event-based multimodal organization converts long demonstrations into ordered subgoal segments and aligns language instructions, visual goals, demonstration context, state transitions, and robot actions within a common training interface.

\subsection{Training Objectives and Three-Stage Curriculum}
\label{sec:training_objective}

WorldScape Policy 2.0 is optimized with a joint video-action objective and, when autonomous planning is enabled, the semantic-forcing objective is further adopted:
\begin{equation}
\mathcal{L} = \mathcal{L}_{act} + \lambda_{w}\mathcal{L}_{world}
+ \lambda_{s}\mathcal{L}_{sem}.
\label{eq:training_objective}
\end{equation}
Here, $\mathcal{L}_{act}$ and $\mathcal{L}_{world}$ are flow-matching objectives for action and future visual-latent generation, respectively, and $\mathcal{L}_{sem}$ follows Eq.~\ref{eq:semantic_forcing}; $\lambda_s$ is 0.001 when semantic forcing is active and zero otherwise. Multimodal conditioning and memory gating are trained end to end through these objectives without separate prompt or memory labels. For embodiment $e$, let $A_0^{(e)}$ denote a clean raw action chunk in $\mathbb{R}^{H\times d_a}$, with $d_a=20$ in the dual-arm setting, and let $A_1^{(e)}\sim\mathcal{N}(0,I)$ be action noise of the same shape. Likewise, $z_0$ and $z_1\sim\mathcal{N}(0,I)$ denote the clean and noisy future visual latents. At $\rho\sim\mathcal{U}(0,1)$, we form $A_{\rho}^{(e)}=(1-\rho)A_0^{(e)}+\rho A_1^{(e)}$ and $z_{\rho}=(1-\rho)z_0+\rho z_1$, with raw-space targets $v^{A,(e)}=A_1^{(e)}-A_0^{(e)}$ and $v^z=z_1-z_0$. Writing $\Xi^{(e)}=\{A_0^{(e)},A_1^{(e)},z_0,z_1\}$, the two flow-matching losses are:
\begin{equation}
\begin{array}{l}
\mathcal{L}_{act}=\mathbb{E}_{e,\rho,\Xi^{(e)}}\!\left[\|\hat v_{\theta}^{A,(e)}(A_{\rho}^{(e)},z_{\rho},\rho)-v^{A,(e)}\|_2^2\right],\\
\mathcal{L}_{world}=\mathbb{E}_{e,\rho,\Xi^{(e)}}\!\left[\|\hat v_{\theta}^{z}(A_{\rho}^{(e)},z_{\rho},\rho)-v^z\|_2^2\right].
\end{array}
\label{eq:act_world_loss}
\end{equation}
Although $E_a^{(e)}$ projects $A_{\rho}^{(e)}$ into DiT tokens for joint video-action attention, the prediction $\hat v_{\theta}^{A,(e)}$ and target $v^{A,(e)}$ remain in raw action space. Thus, $\mathcal{L}_{act}$ does not impose flow matching on action embeddings.

Training follows a three-stage curriculum that progressively acquires multimodal controllability, autonomous planning, and downstream interactive manipulation.

\noindent
\textbf{Stage 1: event-grounded multimodal WAM pretraining.}
We first pretrain the causal video-action backbone on ManipEvent-5M using event-level samples paired with fine-grained captions, goal images, video demonstrations, visual transitions, and robot actions. Different prompt modalities are sampled as interchangeable control conditions: fine-grained captions are injected through T5 cross-attention, while goal images and video demonstrations are encoded as persistent visual prefill. The short-term causal visual memory is already active at this stage: recent observation chunks are dynamically maintained as clean visual prefill, enabling the DiT to model local temporal continuity through causal self-attention. Stage 1 does not yet introduce the VLM-based event memory or autonomous latent subgoal reasoning. Its objective is:
\begin{equation}
\mathcal{L}^{(1)}=\mathcal{L}_{act}+\lambda_w\mathcal{L}_{world}.
\end{equation}
This stage establishes fine-grained language-video-action grounding, short-horizon visual dynamics, and controllable video-action generation before event-level memory reasoning is introduced.

\noindent
\textbf{Stage 2: memory-aware mid-training.}
Starting from the Stage-1 WAM and retaining its short-term causal visual memory, we introduce the VLM planning branch and long short-term event memory for autonomous planning. This stage uses temporally extended robot trajectories, for which episode-level instructions provide global task intent and event-level captions identify the active atomic subgoals. Given only the high-level instruction as input to the planning branch, the model constructs memory-enhanced VLM tokens $\hat{q}_t$ by retrieving global-history, local-active, and event-boundary latents together with the compact history bank; local-active and boundary chunks retain full-token anchor representations for detailed retrieval. The corresponding fine-grained T5 embedding provides a training-time semantic target through $\mathcal{L}_{sem}$, but is not directly used as the autonomous-planning condition. The Stage-2 objective is:
\begin{equation}
\mathcal{L}^{(2)}=\mathcal{L}_{act}+\lambda_w\mathcal{L}_{world}
+\lambda_s\mathcal{L}_{sem}.
\end{equation}
By aligning latent subgoal reasoning with the fine-grained semantic space learned in Stage 1, semantic forcing transfers the pretrained language understanding and controllable video-action generation capabilities to the autonomous planning pathway. The model consequently learns to track task progress, infer the active subgoal, and select the next action without requiring an explicit fine-grained instruction at inference time.

\noindent
\textbf{Stage 3: downstream interactive post-training.}
Finally, we post-train the model on downstream robot tasks using the interaction mode required by each task. Long-horizon manipulation tasks activate high-level instructions, reasoning-augmented event memory, and online visual history for autonomous planning. Fine-grained instruction-following tasks directly condition the WAM on T5 subtask embeddings. In-context adaptation tasks retain goal images or video demonstrations as persistent visual prefill and combine them with online visual memory and, when required, latent reasoning. This stage adapts the shared pretrained model to task-specific embodiments and action distributions while preserving a unified interface for autonomous planning, explicit instruction following, and visual-prompted adaptation.

\section{Experiments}
\label{sec:experiments}

\subsection{Benchmark Setup and Evaluation Protocol}
We evaluate WorldScape Policy 2.0 through simulation benchmarks, controlled ablations, and real-world deployment. For the main RoboTwin 2.0 comparison~\cite{chen2025robotwin}, all methods are fine-tuned for 50K steps on the full clean-plus-randomized training data to match the setting used by the compared baselines, and are evaluated over 100 trials per task across 50 tasks under both clean and randomized conditions. Across these settings, we assess task success, subgoal completion, instruction following, progress consistency, and in-context adaptation.

Controlled ablations on RoboTwin 2.0 isolate the contributions of memory components, staged training and semantic forcing. We further evaluate the model on an AgileX Robotics platform across four main capabilities: long-horizon autonomous planning, memory-dependent reasoning, cross-embodiment ICL (In-Context Learning), and sequential fine-grained control. For each real-world task, we perform 20 trials and report the average success rate.

\subsection{Implementation Details}

\noindent
\textbf{Model configurations.}
For event annotation, we use Qwen3-VL-32B~\cite{Qwen3-VL} to generate and verify segment-level subtask captions, temporal boundaries, and prompt metadata. The reasoning-augmented LSTM uses a lightweight Qwen3-VL-4B model~\cite{Qwen3-VL} as the latent reasoning backbone. To reduce memory cost, this module receives only the head-view observation image, resized to $320 \times 160$. For each action chunk, a single VLM prefill jointly encodes this image and the task-conditioned planning prompt, after which the model autoregressively generates $K=4$ planning tokens from the same cached context.

The WAM input concatenates three camera views into a single visual canvas. The video DiT is initialized from the Wan2.2-5B text-to-video pretrained model~\cite{wan2025wan}, and the video and action DiTs share the same backbone. Video-action interaction uses joint bidirectional attention between video and action tokens, while attention to clean tokens and visual-history tokens remains causal to preserve autoregressive prediction. During multi-embodiment pretraining, each embodiment uses its own action encoder--decoder pair $(E_a^{(e)},D_a^{(e)})$, while the video-action backbone is shared. The short-term visual memory retains up to 4 recent chunks. The long-term memory keeps 8 history chunks by default, and this length can be adjusted according to task horizon and memory demand.

\noindent
\textbf{Training setup.}
The action interface follows Sec.~\ref{sec:problem_formulation}: each arm contributes a chunk-relative 3D delta position, a continuous 6D relative-rotation representation, and an absolute one-DoF gripper command, giving a 20D raw action for dual-arm control. As specified in Eq.~\ref{eq:act_world_loss}, action flow matching is optimized directly in this raw space; the action encoder and decoder serve only as embodiment-specific interfaces to the shared DiT. Event-grounded pretraining uses batch size 768 and learning rate $5\times10^{-4}$; post-training uses batch size 128 and learning rate $6\times10^{-5}$ for 50K steps.

\begin{table}[tb!]
\footnotesize
\begin{center}
\caption{\textbf{Evaluation on the RoboTwin 2.0 benchmark.} Results are averaged across all 50 tasks under clean and randomized evaluation settings; all methods use clean-plus-randomized training data for a consistent comparison.}
\label{tab:robotwin_compare}
\begin{tabular*}{0.7\columnwidth}{@{}l@{\extracolsep{\fill}}ccc@{}}
\toprule[1pt]
\textbf{Model} & \multicolumn{1}{c}{\textbf{Clean}} & \multicolumn{1}{c}{\textbf{Randomized}} & \multicolumn{1}{c}{\textbf{Average}} \\ \midrule
\multicolumn{4}{c}{\cellcolor[HTML]{DADADA}\texttt{\# Vision-Language-Action (VLA) Models}} \\ \midrule

  $\pi_{0}$~\cite{black2024pi_0} & 65.9\% & 58.4\% & 62.2\% \\ 
  X-VLA~\cite{zheng2025x} & 72.9\% & 72.8\% & 72.9\% \\ 
  $\pi_{0.5}$~\cite{Pi-0.5} & 82.7\% & 76.8\%  & 79.8\% \\ 
  Abot-M0~\cite{chen2026abot} & 81.2\% & 80.4\% & 80.8\% \\ 
  LingBot-VLA~\cite{wu2026pragmatic} & 86.5\% & 85.3\% & 85.9\% \\ 
  HoloBrain-0-QW~\cite{lin2026holobrain} & 91.9\% & 92.3\% & 92.1\% \\ 
\midrule
\multicolumn{4}{c}{\cellcolor[HTML]{DADADA}\texttt{\# World Action Models}} \\ \midrule
  Motus~\cite{bi2025motus} & 88.7\% & 87.0\% & 87.9\% \\ 
  GigaWorld-Policy~\cite{ye2026gigaworld} & 85.6\% & 85.3\% & 85.5\% \\ 
  LingBot-VA~\cite{li2026causal} & 92.9\% & 91.6\% & 92.3\% \\ 
  Fast-WAM~\cite{yuan2026fast} & 91.9\% & 91.8\% & 91.9\% \\ 
  Abot-M0.5~\cite{chen2026abot} & 94.0\% & \textbf{94.2\%} & 94.1\% \\ 
  LingBot-VA 2.0~\cite{zhang2026native} & 93.8\% & 93.4\% & 93.6\% \\ 
  WorldScape Policy 1.0~\cite{su2026worldscape} &  93.2\% &  91.7\% & 92.5\% \\

  \textbf{WorldScape Policy 2.0} &  \textbf{94.3\%} &  \textbf{94.2\%} & \textbf{94.3\%} \\
  \bottomrule[1pt]
\end{tabular*}
\end{center}
\end{table}

\subsection{Simulation Benchmark Results}



We evaluate manipulation robustness across all 50 tasks of RoboTwin 2.0~\cite{chen2025robotwin} under both clean and randomized settings. The benchmark covers diverse bimanual skills and introduces visual and physical perturbations in the randomized setting, measuring robustness under the same clean-plus-randomized training protocol used for all methods in Table~\ref{tab:robotwin_compare}.

As shown in Table~\ref{tab:robotwin_compare}, WorldScape Policy 2.0 achieves the best overall average success rate of 94.3\%, with 94.3\% under clean conditions and 94.2\% under randomization. It outperforms the representative VLA baseline $\pi_{0.5}$~\cite{Pi-0.5} by 11.6\%, 17.4\%, and 14.5\% on clean, randomized, and average success, respectively. Compared with Fast-WAM~\cite{yuan2026fast}, a representative WAM baseline, it still improves clean, randomized, and average success by 2.4\% each. While compared with LingBot-VA 2.0~\cite{zhang2026native}, a recent natively pretrained WAM, our method further improves clean, randomized, and average success by 0.5\%, 0.8\%, and 0.7\%, respectively, further validating the effectiveness of our event-grounded pretraining strategy. The negligible gap between clean and randomized performance further demonstrates the robustness and strong generalization capability of WorldScape Policy 2.0 across diverse bimanual manipulation tasks.

\begin{figure*}[t]
\centering
\includegraphics[width=0.95\textwidth]{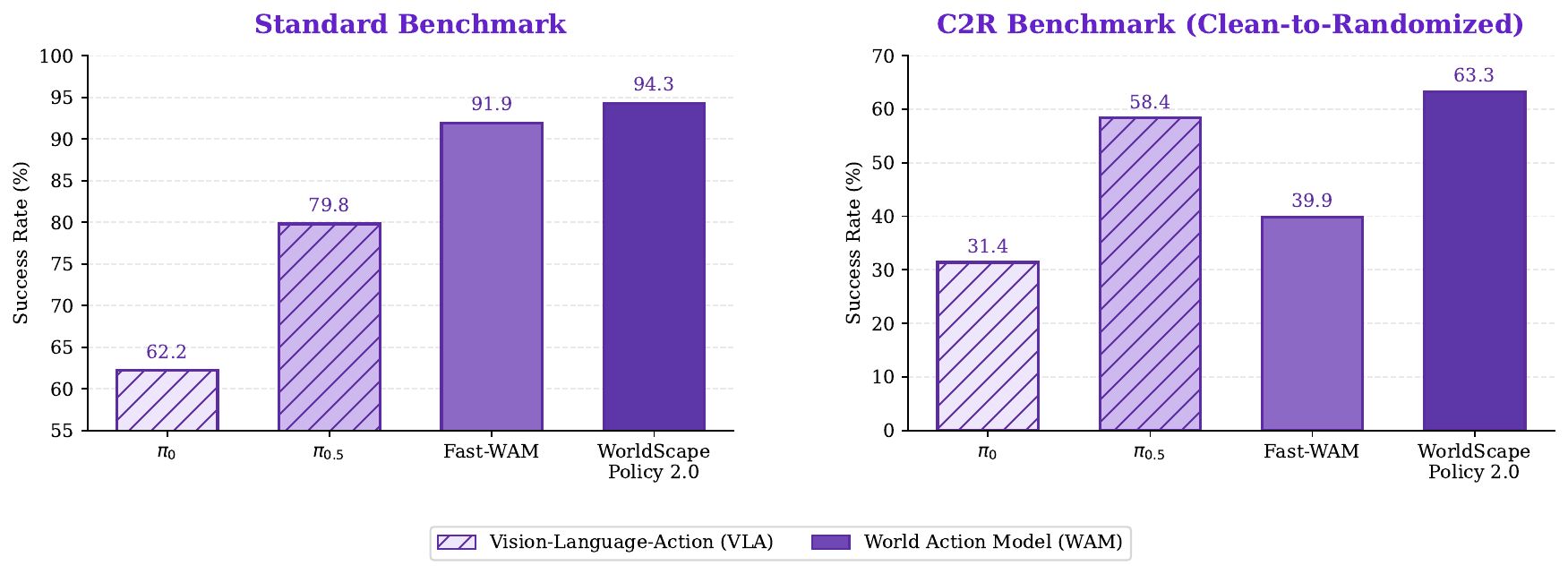}
\caption{\textbf{Comparison on the RoboTwin 2.0 standard and C2R benchmarks.} The standard benchmark uses clean-plus-randomized training data, whereas C2R trains policies only on clean demonstrations and reports the average success rate across clean and randomized evaluation settings. Hatched bars denote VLA methods, and solid bars denote WAM methods.}
\label{fig:robotwin_standard_c2r}
\end{figure*}

Figure~\ref{fig:robotwin_standard_c2r} further compares the standard benchmark with the stricter Clean-to-Randomized (C2R) setting. Unlike the standard protocol, C2R uses only clean demonstrations for training and evaluates the resulting policy under both clean and randomized conditions; the reported score averages success across these two evaluation settings, providing a balanced measure of in-domain performance and OOD generalization without exposure to randomized training data. Under this setting, WorldScape Policy 2.0 achieves an average success rate of 63.3\%, outperforming $\pi_0$, $\pi_{0.5}$, and Fast-WAM by 31.9, 4.9, and 23.4 percentage points, respectively.

\begin{figure*}[t]
  \centering
  \includegraphics[width=1.0\textwidth]{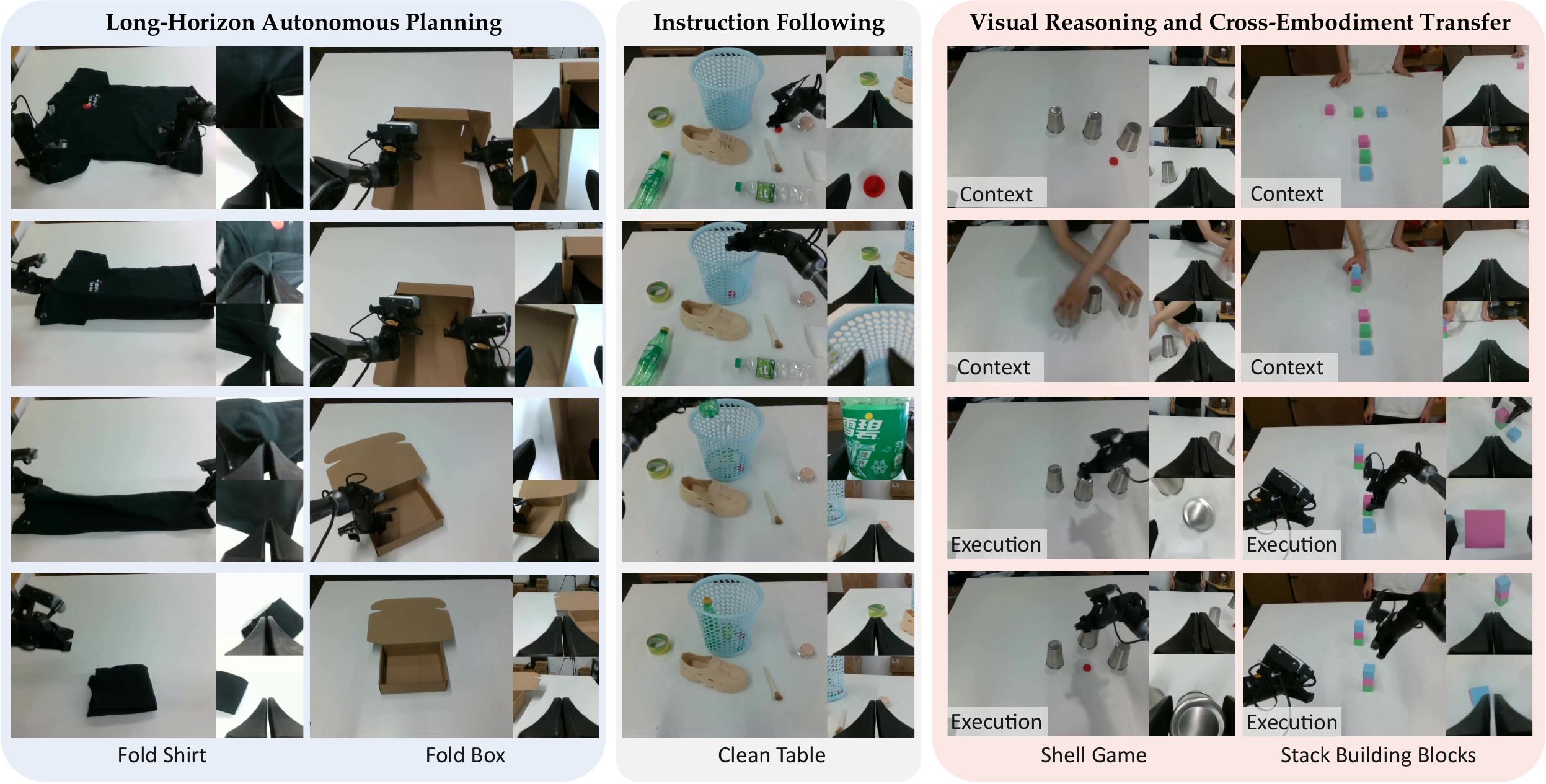}
  \caption{\textbf{Illustration of real-world tasks on the dual-arm PiPER platform.} The benchmark covers long-horizon autonomous planning through \texttt{shirt and box folding}, fine-grained instruction following through \texttt{table cleaning}, and visual-context reasoning and cross-embodiment transfer through the \texttt{shell game} and demonstration-conditioned \texttt{block stacking}, respectively.}
  \label{fig:realworld_tasks}
\end{figure*}

\begin{table*}[t]\footnotesize
\begin{center}
\caption{\textbf{Real-robot manipulation benchmark across diverse capabilities.} The evaluation reports episode-level task success rate for autonomous planning, sequential instruction following, and in-context reasoning / skill transfer with text, goal-image, and video-demonstration prompts.}
\label{tab:real_robot}
\setlength{\tabcolsep}{0.45cm}
\resizebox{1\textwidth}{!}{
\begin{tabular}{c|c|c|ccc}
\toprule[1pt]
\textbf{Capability} & \textbf{Tasks} & \textbf{Prompt / Context} & $\pi_{0.5}$  & DreamZero & \textbf{\makecell[c]{WorldScape \\ Policy 2.0}} \\
\midrule
\multirow{2}{*}{\makecell[c]{Long-Horizon\\Autonomous Planning}} & \texttt{Folding Clothes} & \multirow{2}{*}{\makecell[c]{Global Instruction}} & 60\% & 45\% & \textbf{75\%} \\
& \texttt{Folding Box} & & 65\% & 55\% & \textbf{75\%} \\
\midrule
\makecell[c]{Memory-Dependent\\Visual Reasoning} & \texttt{Shell Game} & Video Demo & 30\% & 50\% & \textbf{75\%} \\
\midrule
\multirow{2}{*}{\makecell[c]{Cross-Embodiment \\ Skill Transfer}} & \multirow{2}{*}{\makecell[c]{\texttt{Stacking Blocks}}} & Goal Image & 10\% & 20\% & \textbf{60\%} \\
& & Video Demo & 20\% & 20\% & \textbf{70\%} \\
\midrule
\makecell[c]{Sequential\\Fine-Grained Control} & \makecell[c]{\texttt{Clean Table}\\(Full Sequence)} & \makecell[c]{Sequential\\Subtask Captions} & 70\% & 60\% & \textbf{80\%} \\
\bottomrule[1pt]
\end{tabular}}
\end{center}
\end{table*}

\begin{table*}[t]\footnotesize
\begin{center}
\caption{\textbf{Fine-grained text-instruction following on the real-world} \texttt{table cleaning} \textbf{task.} In-domain object categories are included in task-specific post-training, whereas out-of-domain categories are held out and introduced only at evaluation. Each row independently evaluates one atomic instruction under a reset scene. We report instruction-level success rate over 20 trials.}
\vspace{-0.2cm}
\label{tab:fine_grained_instruction}
\setlength{\tabcolsep}{0.28cm}
\renewcommand{\arraystretch}{1.12}
\resizebox{\textwidth}{!}{
\begin{tabular}{c|c|l|ccc}
\toprule[1pt]
\textbf{Task} & \textbf{Domain Split} & \textbf{Fine-Grained Text Instruction} & $\pi_{0.5}$ & DreamZero & \textbf{\makecell[c]{WorldScape \\ Policy 2.0}} \\
\midrule
\multirow{6}{*}{\texttt{Clean Table}}
& \multirow{3}{*}{In-domain}
& Pick up the \emph{white tissue} and place it into the \emph{basket}. & 70\% & 75\% & \textbf{80\%} \\
& & Pick up the \emph{paper cup} and place it into the \emph{basket}. & \textbf{90\%} & 75\% & \textbf{90\%} \\
& & Pick up the \emph{plastic bottle} and place it into the \emph{basket}. & \textbf{90\%} & 70\% & \textbf{90\%} \\
\cmidrule(lr){2-6}
& \multirow{3}{*}{Out-of-domain}
& Pick up the \emph{black pen} and place it into the \emph{basket}. & 50\% & 40\% & \textbf{70\%} \\
& & Pick up the \emph{green tape} and place it into the \emph{basket}. & 55\% & 35\% & \textbf{60\%} \\
& & Pick up the \emph{beige shoe} and place it into the \emph{basket}. & 25\% & 15\% & \textbf{50\%} \\
\midrule
\multicolumn{3}{c|}{\textbf{In-Domain Average}} & 83.3\% & 73.3\% & \textbf{86.7\%} \\
\multicolumn{3}{c|}{\textbf{Out-of-Domain Average}} & 43.3\% & 30.0\% & \textbf{60.0\%} \\
\multicolumn{3}{c|}{\textbf{Overall Average}} & 63.3\% & 51.7\% & \textbf{73.3\%} \\
\bottomrule[1pt]
\end{tabular}}
\end{center}
\end{table*}

\subsection{Real-World Evaluation Results}
We evaluate five real-world tasks on the dual-arm PiPER platform, as illustrated in Fig.~\ref{fig:realworld_tasks}. Table~\ref{tab:real_robot} covers comprehensive evaluation across four core capabilities: \ding{182} \textit{\textbf{Long-horizon autonomous planning}} evaluates the success rates of \texttt{clothes and box folding} tasks respectively under a global instruction. \ding{183} \textit{\textbf{Memory-dependent visual reasoning}} uses a video-conditioned \texttt{shell game} task for evaluation, requiring the robot to track history and infer a hidden object state. \ding{184} \textit{\textbf{Cross-embodiment skill transfer}} evaluates \texttt{block stacking} policy conditioned on either a goal image or a demonstration video. \ding{185} \textit{\textbf{Sequential fine-grained controllability}} evaluates the full \texttt{table-cleaning} sequence driven by successive subtask captions. Table~\ref{tab:fine_grained_instruction} further isolates text grounding through six atomic \texttt{table-cleaning} instructions under reset scenes. As discussed in the Sec.~\ref{sec:introduction}, standard VLA policies such as $\pi_{0.5}$~\cite{Pi-0.5} rely on static single-frame observations as input and therefore lack the temporal context required by the \texttt{shell game} task and demonstration-conditioned \texttt{block stacking} task. For a fair and task-feasible comparison, we temporally extend the input of $\pi_{0.5}$ in these evaluations, allowing it to access the observation history necessary to attempt the tasks.

Across these real-world tasks, WorldScape Policy 2.0 outperforms both the representative VLA method $\pi_{0.5}$~\cite{Pi-0.5} and the WAM method DreamZero~\cite{ye2026world} in overall episode success rate. As shown in Table~\ref{tab:real_robot}, it achieves 75\% success on both long-horizon \texttt{folding} tasks and 80\% on sequential \texttt{table cleaning}, consistently exceeding both baselines. The advantage is even more pronounced on visual-context tasks, where it reaches 75\% on the \texttt{shell game} and 60--70\% on cross-embodiment \texttt{block stacking}, demonstrating strong memory-dependent reasoning and visual-prompt transfer capabilities. The fine-grained instruction-following results in Table~\ref{tab:fine_grained_instruction} further validate the benefit of the proposed event-grounded pretraining. WorldScape Policy 2.0 achieves the best in-domain average success rate (86.7\%) and a substantially higher overall average (73.3\%) than $\pi_{0.5}$ (63.3\%) and DreamZero (51.7\%). The improvement is especially clear under held-out object categories: WorldScape Policy 2.0 reaches 60.0\% OOD (Out-of-Domain) success rate, compared with 43.3\% for $\pi_{0.5}$ and 30.0\% for DreamZero. These results indicate that fine-grained event-level pretraining improves atomic instruction grounding and compositional generalization, demonstrating the effectiveness and necessity of the proposed framework for controllable manipulation.



\subsection{Ablation Study}
Unless otherwise specified, all ablative experiments are conducted on the RoboTwin 2.0 benchmark and follow the official clean-only training setting: each variant is trained only on the clean data split, rather than the full clean-plus-randomized data used for the main comparison in Table~\ref{tab:robotwin_compare}. Consequently, the absolute ablation scores should not be directly compared with the headline results in Table~\ref{tab:robotwin_compare}.

\begin{table}[t]\footnotesize
\begin{center}
\caption{\textbf{Component analysis on RoboTwin 2.0}. STM denotes short-term visual memory, LTM denotes long-term event memory, and LSR denotes latent subgoal reasoning.}
\vspace{-0.2cm}
\label{tab:memory_ablation}
\setlength{\tabcolsep}{0.8cm}
\begin{tabular}{ccc|ccc}
\toprule[1pt]
\textbf{STM} & \textbf{LTM} & \textbf{LSR} & \textbf{Clean}  & \textbf{Randomized} & \textbf{Average}  \\
\midrule
\ding{55} & \ding{55} & \ding{55} & 78.86\% & 29.74\% & 54.30\% \\ 
\checkmark & \ding{55} & \ding{55} & 84.92\% & 38.50\% & 61.71\% \\
\checkmark & \checkmark & \ding{55} & 84.74\% & 40.02\% & 62.38\% \\
\checkmark & \checkmark & \checkmark & \textbf{85.04\%} & \textbf{41.58\%} & \textbf{63.31\%} \\
\bottomrule[1pt]
\end{tabular}
\end{center}
\end{table}



\begin{table}[t]\footnotesize
\begin{center}
\caption{\textbf{Ablation of Stage-1 event-grounded pretraining and Stage-2 memory-aware mid-training on RoboTwin 2.0.} SF denotes semantic forcing. All variants use the same Stage-3 downstream post-training protocol.}
\vspace{-0.2cm}
\label{tab:pretrain_ablation}
\setlength{\tabcolsep}{0.8cm}
\begin{tabular}{ccc|ccc}
\toprule[1pt]
\textbf{Stage-1} & \textbf{Stage-2} & \textbf{SF} & \textbf{Clean} & \textbf{Randomized} & \textbf{Average} \\
\midrule
\ding{55} & \ding{55} & \ding{55} & 75.38\% & 28.06\% & 51.72\% \\
\checkmark & \ding{55} & \ding{55} & 81.90\% & 37.34\% & 59.62\% \\
\checkmark & \checkmark & \ding{55} & 84.30\% & 40.56\% & 62.43\% \\
\checkmark & \checkmark & \checkmark & \textbf{85.04\%} & \textbf{41.58\%} & \textbf{63.31\%} \\
\bottomrule[1pt]
\end{tabular}
\end{center}
\end{table}


\ding{182} \textit{\textbf{Memory and latent reasoning components.}}
Table~\ref{tab:memory_ablation} progressively adds short-term visual memory (STM), long-term event memory (LTM), and latent subgoal reasoning (LSR). Compared with the no-memory baseline, STM improves clean, randomized, and average success by 6.06\%, 8.76\%, and 7.41\%, respectively. Adding LTM yields 84.74\%, 40.02\%, and 62.38\% on the three metrics, substantially improving randomized and average success while maintaining comparable clean performance, which shows that event-level progress memory complements recent visual context. Finally, with the help of implicit subgoal reasoning, WorldScape Policy 2.0 achieves the best results as expected. Overall, the complete model consistently outperforms the no-memory baseline under both clean and randomized settings, strongly validating the importance of integrating short-term visual memory, long-term event memory, and latent subgoal planning for robust manipulation.

\ding{183} \textit{\textbf{Training stages and semantic forcing.}} Table~\ref{tab:pretrain_ablation} separates the effects of Stage-1 
event-grounded pretraining, Stage-2 memory-aware mid-training, and 
semantic forcing (SF), while keeping Stage-3 downstream 
post-training fixed. The four rows compare training from scratch, 
Stage 1 alone, Stages 1--2 without SF, and the complete curriculum 
with SF. This design measures whether multimodal event-grounded pretraining provides a useful initialization, whether memory mid-training adds progress-aware prediction, and whether semantic forcing transfers explicit event semantics into latent subgoal reasoning. It can be observed that Stage-1 pretraining yields the largest improvement, particularly under randomization, while Stage-2 mid-training and semantic forcing provide further consistent gains through progress-aware memory and semantically aligned latent reasoning. The monotonic improvement across stages validates the effectiveness and complementarity of the proposed multi-stage pretraining framework for robust manipulation.

\section{Conclusion}
\label{sec:conclusion}

We presented WorldScape Policy 2.0, a controllable World Action Model for long-horizon robotic manipulation. Its reasoning-augmented long short-term memory couples causal short-term visual context with event-level progress memory and latent subgoal reasoning, enabling a single video-action model to support autonomous planning from high-level instructions, fine-grained text control, goal-image conditioning, and demonstration-based in-context adaptation. To train these capabilities, we introduced ManipEvent-5M, an event-grounded multimodal dataset with nearly five million segments, and a three-stage curriculum that transfers explicit event semantics into autonomous reasoning through semantic forcing. Evaluations on both simulation and real-world platforms demonstrate the complementary roles of visual memory, event memory, and latent reasoning, substantially outperforming representative VLA and WAM baselines and validating the effectiveness of event-level pretraining. These results advance WAMs from passive future prediction toward memory-grounded and multimodally controllable manipulation.

\clearpage
\newpage
\bibliography{ref}

\end{document}